\documentclass[final]{cvpr}
\usepackage{times}
\usepackage{epsfig}
\usepackage{graphicx}
\usepackage{amsmath}
\usepackage{amssymb}
\usepackage{csquotes}
\usepackage{rotating}
\usepackage{color}
\usepackage{algorithm}
\usepackage{algorithmicx}
\usepackage{algpseudocode}
\usepackage{subfigure}
\usepackage{arydshln}
\usepackage{multicol}
\usepackage{wrapfig}
\usepackage{multirow}
\usepackage{booktabs}
\usepackage{bbding}
\usepackage[table,xcdraw]{xcolor}
\usepackage[normalem]{ulem}

\usepackage[pagebackref=true,breaklinks=true,colorlinks,bookmarks=false]{hyperref}

\def\cvprPaperID{771} 
\def\confYear{CVPR 2021}

\begin{document}

\title{Uncertainty-aware Joint Salient Object and Camouflaged Object Detection}


\author{
Aixuan Li$^{1,\ddag}$~
Jing Zhang$^{2,3,\ddag}$~
Yunqiu Lv$^{1}$~
Bowen Liu$^{1}$~ 
Tong Zhang$^{4}$~ 
Yuchao Dai$^1$\href{mailto:daiyuchao@gmail.com}{\Envelope}~
\\
$^1$ Northwestern Polytechnical University, China \quad
$^2$ Australian National University, Australia\\
$^3$ CSIRO, Australia \quad
$^4$ EPFL, Switzerland\\
{\tt \small $\ddag$ Equal contributions; \Envelope~Corresponding author:daiyuchao@nwpu.edu.cn
}
}

\maketitle

\begin{abstract}

Visual salient object detection (SOD) aims at finding the salient object(s) that attract human attention, while camouflaged object detection (COD) on the contrary intends to discover the camouflaged object(s) that hidden in the surrounding. In this paper, we propose a paradigm of leveraging the contradictory information to enhance the detection ability of both salient object detection and camouflaged object detection. We start by exploiting the easy positive samples in the COD dataset to serve as hard positive samples in the SOD task to improve the robustness of the SOD model. Then, we introduce a \enquote{similarity measure} module to explicitly model the contradicting attributes of these two tasks.
Furthermore, considering the uncertainty of labeling in both tasks’ datasets, we propose an adversarial learning network to achieve both higher order similarity measure and network confidence estimation. Experimental results on benchmark datasets demonstrate that our solution leads to state-of-the-art (SOTA) performance for both tasks\footnote{Our code is publicly available at: \url{https://github.com/JingZhang617/Joint_COD_SOD}}.

\end{abstract}

\section{Introduction}
\begin{figure}[tp]
   \begin{center}
   \begin{tabular}{c@{ }}
   {\includegraphics[width=0.80\linewidth]{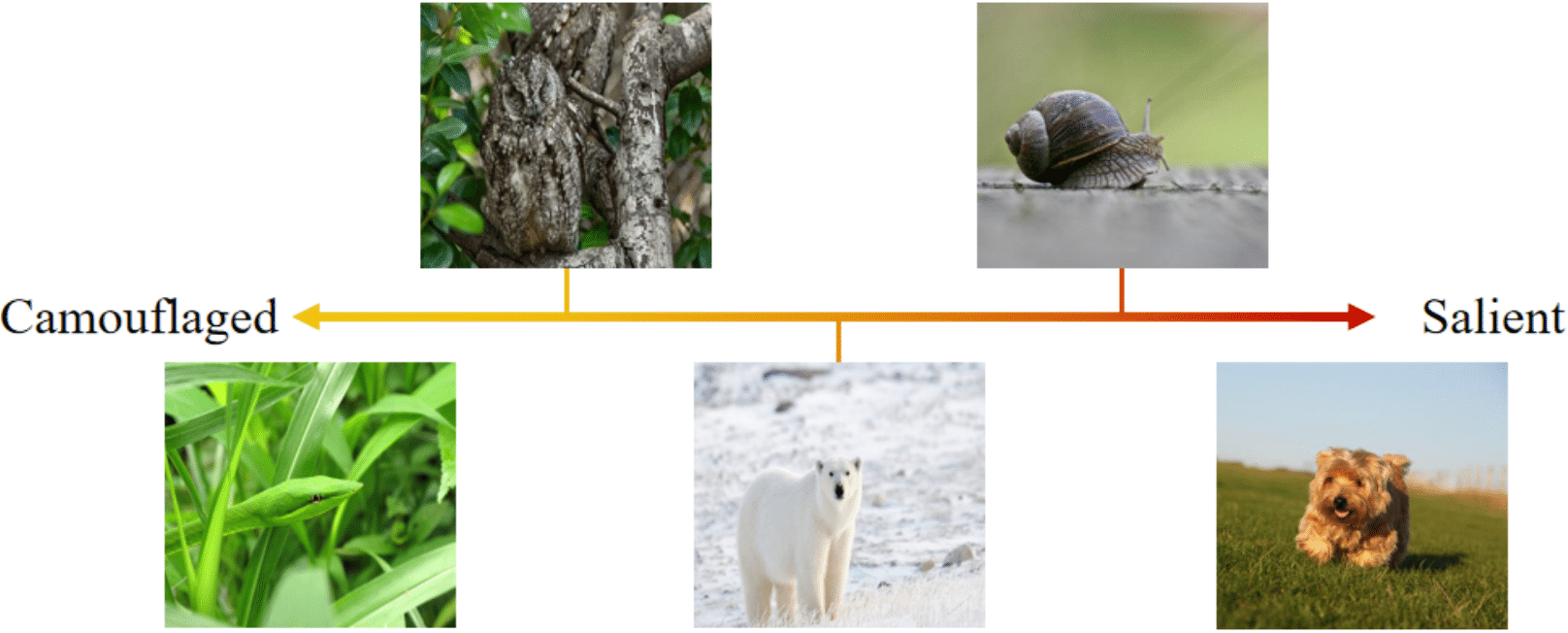}}
    \\
   \end{tabular}
   \end{center}
   \caption{Illustration of the
   transition from camouflaged objects to salient objects, where the image
   in the middle could belong to both camouflaged object dataset and salient object dataset.
   \vspace{-2mm}
   }
   
   
\label{fig:relationship of sod and cod}
\end{figure}

Visual salient object detection (SOD) aims to localize the most salient region(s) of the images that attract human attention. To be qualified as a \enquote{salient} object, one should have high contrast compared with its global and local context. The camouflaged objects oppositely usually share similar structure or texture information with the environment, which try hard to fade themselves into the local context. In this way, the SOD models \cite{scrn_sal,wei2020f3net,wang2020progressive,basnet_sal} are designed based on both global contrast and local contrast, while the COD models \cite{fan2020camouflaged,le2019anabranch,mei2021Ming,zhai2021Mutual,yunqiu_cod21} usually avoid searching the camouflaged objects in those salient regions.
We notice that a higher level of saliency indicates a lower level of camouflage and vice versa as shown in Fig.~\ref{fig:relationship of sod and cod}. This observation shows that the salient objects and camouflaged objects are two contradicting categories of objects.
However, there still exist objects that are both salient and camouflaged, \eg the polar bear in the middle of Fig.~\ref{fig:relationship of sod and cod}, which indicates that these two tasks are partially positively related at the dataset level.

Existing SOD models \cite{scrn_sal,cpd_sal,zhou2020interactive,wei2020f3net,basnet_sal} mainly focus on two directions: 1) building effective saliency network \cite{cpd_sal,zhou2020interactive} for accurate saliency detection with pixel-wise accuracy constraint; and 2) designing appropriate loss functions \cite{wei2020f3net,basnet_sal} to achieve structure-preserving saliency detection. 
The former digs into network structure, while the latter cares more about network loss function. We argue that an effective training dataset can lead to more performance gain in addition to network structure design or loss function, it's the training data that is regressed. 


One typical solution to explore the training dataset is data augmentation, which usually involves linear or non-linear transformation of the dataset. We find that, although performance improvement can be obtained with some basic data augmentation, \eg image flipping, rotation, cropping, and \etc, none of these methods are specially designed for saliency detection. As a context-based task, a more effective data augmentation technique should be
context-aware.
For SOD, the salient objects are those that can be easily detected, or the high-contrast objects as shown in Fig.~\ref{fig:relationship of sod and cod}. We intend to augment the dataset to include lower contrast samples.
Considering the partial positively related attribute of SOD and COD at dataset level, we intend to design a joint learning framework to learn both tasks and select
easy samples from COD (\eg the polar bear) as hard samples for SOD, achieving contrast-level data augmentation.

\begin{figure*}[!htp]
   \begin{center}
   \begin{tabular}{c@{ }}
   {\includegraphics[width=0.75\linewidth]{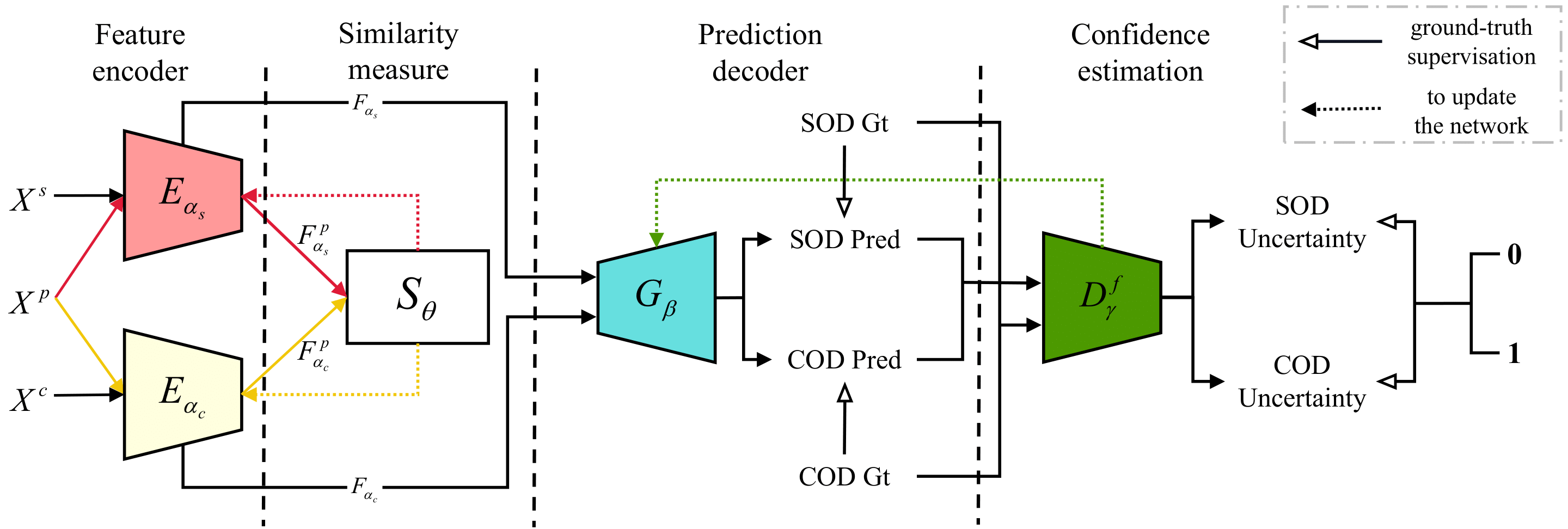}}
    \\
   \end{tabular}
   \end{center}
   \caption{Overview of the proposed network. The \enquote{Feature encoder} module extracts task-specific features for image $X^s$ and $X^c$ from SOD and COD dataset respectively. And for the connection modeling data $X^p$, we introduce \enquote{Similarity measure} to explicitly model the contradicting attribute of SOD and COD.
   The shared \enquote{Prediction decoder} module is used to generate predictions for both tasks.
   The shared \enquote{Confidence estimation} module is a fully convolutional discriminator, which estimates the pixel-wise confidence of network prediction.  
   }
  \vspace{-2mm}
\label{fig:network_overview}
\end{figure*}

Joint training is mostly designed for positively related tasks \cite{wang2016joint,fu2020jl,zeng2019joint}. In contrary, we propose to integrate two contradicting tasks (SOD and COD) into one network with 
a \enquote{Similarity measure} module as shown in Fig.~\ref{fig:network_overview}. 
The basic assumption of our similarity measure module is that the activated regions of the same image for the two tasks should be different, leading to latent features apart from each other. 
To this end, we introduce the third dataset, \eg PASCAL VOC 2007 images \cite{everingham2007pascal} in particular, to our framework serving as the connection modeling dataset.
The goal of these extra images is to achieve similarity measures and force the two tasks to focus on different regions of the image.

Moreover, as shown in Fig.~\ref{fig:relationship of sod and cod}, the salient object is salient in both local and global contexts, while the camouflaged object is hiding in its local context.
Due to the high contrast, the local context of the salient objects is easier to model than that of the camouflaged objects, as there exists no clear boundary between camouflaged objects and their surrounding. By jointly training a salient object detection network and camouflaged object detection network, the salient object branch can learn precise local context information for accurate camouflaged object detection. 


Lastly, we observe two types of uncertainty while labeling the dataset for each task. For salient object detection, the subjective nature of saliency \cite{ucnet_sal,zhang2020uncertainty,zhang2020learning} leads to ambiguity of prediction, as shown in Fig.~\ref{fig:labeling_uncertainty}(a). For camouflaged object detection, the uncertainty comes from the difficulty in fully annotating the camouflaged objects as they usually share similar color or texture with the environment, as shown in Fig.~\ref{fig:labeling_uncertainty}(c). We then introduce adversarial training to explicitly model the confidence of network predictions, and estimate model uncertainty for both tasks.


We summarize our main contributions as: 1) We introduce the first joint salient object detection and camouflaged object detection network within an adversarial learning framework to explicitly model prediction uncertainty of each task. 2) We design the similarity measure module to explicitly model the \enquote{contradicting} attributes of the two tasks. 3) We present a data interaction strategy and treat easy samples from camouflage dataset as hard samples for saliency detection, achieving a robust saliency model.

\section{Related Work}

\noindent\textbf{Salient Object Detection Models} 
Existing deep saliency detection models \cite{cpd_sal,scrn_sal,wang2020progressive,Liu19PoolNet,wei2020f3net,feng2019attentive,basnet_sal} are mainly designed to achieve structure-preserving saliency predictions. \cite{basnet_sal,scrn_sal} introduced auxiliary edge detection branch to produce a saliency map with precise structure information. Wei \etal \cite{wei2020f3net} presented structure-aware loss function to penalize prediction along object edges. Wu \etal \cite{cpd_sal} designed a cascade partial decoder to achieve accurate saliency detection with finer detailed information. Feng \etal \cite{feng2019attentive} proposed a boundary-aware mechanism to improve the accuracy of network prediction on the boundary. There also exist salient object detection models that benefit from data of other sources. \cite{wang2018salient,saliency_unified} integrated fixation prediction and salient object detection in a unified framework to explore the connections of the two related tasks. Zeng \etal \cite{zeng2019joint} presented to jointly learn a weakly supervised semantic segmentation and fully supervised salient object detection model to benefit from both tasks.

\noindent\textbf{Camouflaged Object Detection Models} Camouflage models are designed to discover the camouflaged object(s) hidden in the surrounding. Different from salient objects, which are those attracting human attention, camouflaged objects are those trying to decrease the conspicuousness.
The concept of camouflage is usually associated with context \cite{behrens1988theories,behrens2018seeing,cuthill2019camouflage}.
Cuthill \etal \cite{cuthill2005disruptive} presented that an effective camouflage includes two mechanisms: the background pattern matching, where the color is similar to the environment, and the disruptive coloration, which usually involves bright colors along edge, and makes the boundary between camouflaged objects and the background unnoticeable.
Bhajantri \etal \cite{bhajantri2006camouflage} utilized co-occurrence matrix to detect defective. Pike \etal \cite{pike2018quantifying} combined several salient visual features to quantify camouflage, which could simulate the visual mechanism of a predator.
In the field of deep learning, Fan \etal \cite{fan2020camouflaged} proposed the first publicly available camouflage deep network with the largest camouflaged object training set.

\noindent\textbf{Multi-task Learning} 
The basic assumption behind multi-task learning is that there exists shared information among different tasks. In this way, multi-task learning is widely used to extract complementary information about positively related tasks. Kalogeiton \etal \cite{kalogeiton2017joint} jointly detected objects and actions in a video scene.
Zhen \etal \cite{zhen2020joint} designed a joint semantic segmentation and boundary detection framework by iteratively fusing feature maps generated for each task with pyramid context module.
In order to solve the problem of insufficient supervision in semantic alignment and object landmark detection, Jeon \etal \cite{jeon2019joint} designed a joint loss function to impose constraints between tasks, and only reliable matched pairs were used to improve the model robustness with weak supervision.
Joung \etal \cite{joung2020cylindrical} solved the problem of object viewpoint changes in 3D object detection and viewpoint estimation with a cylindrical convolutional network, which obtains view-speciﬁc features with structural information at each viewpoint for both two tasks.
Luo \etal \cite{luo2020multi} presented a multitask framework for referring expression comprehension and segmentation.

\begin{figure}[tp]
   \begin{center}
   \begin{tabular}{{c@{ } c@{ } c@{ } c@{ }}}
   {\includegraphics[height=0.16\linewidth]{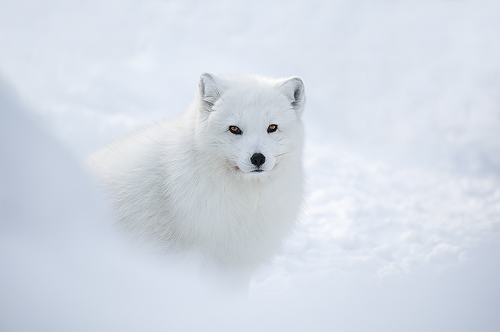}}&
    {\includegraphics[height=0.16\linewidth]{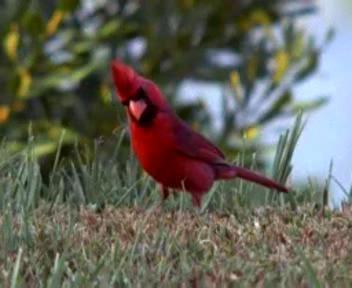}}&
    {\includegraphics[height=0.16\linewidth]{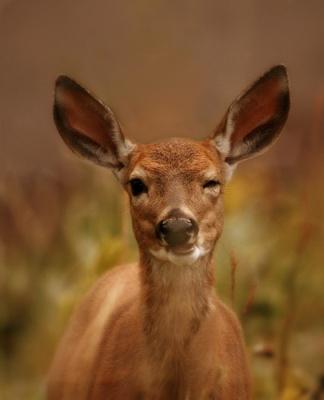}}&
    {\includegraphics[height=0.16\linewidth]{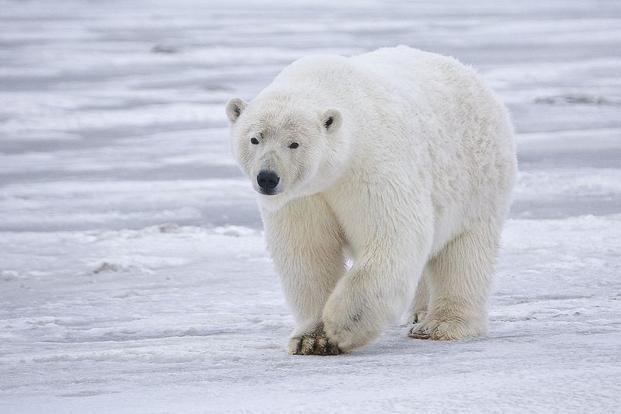}} \\
   {\includegraphics[height=0.16\linewidth]{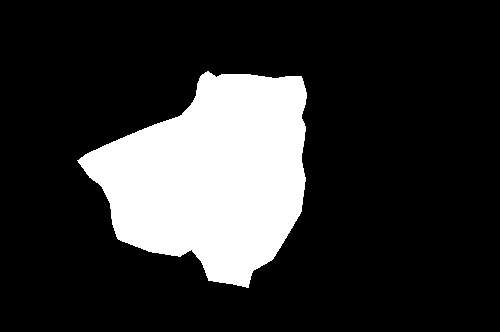}}&
    {\includegraphics[height=0.16\linewidth]{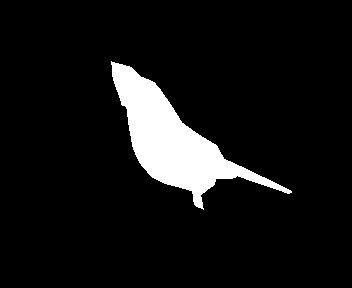}}&
    {\includegraphics[height=0.16\linewidth]{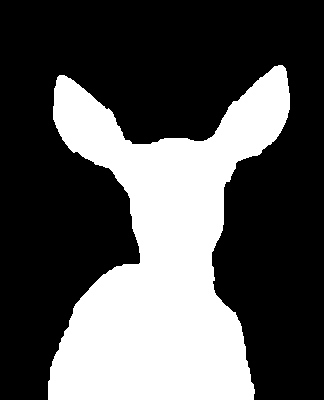}}&
    {\includegraphics[height=0.16\linewidth]{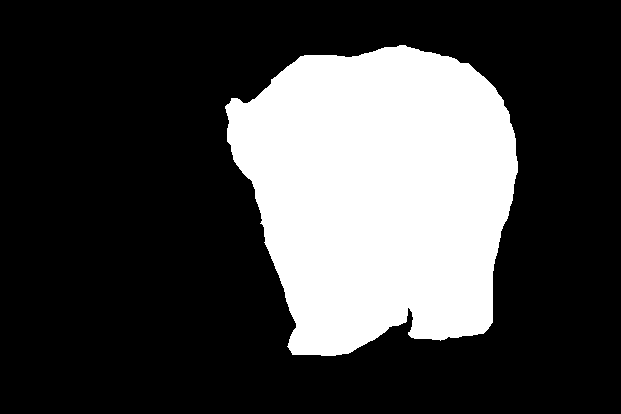}} \\
   \end{tabular}
   \end{center}
    \caption{Selected easy samples from the COD training dataset.} 
    \vspace{-2mm}
    \label{fig:selected_easy_samples_from_cod}
\end{figure}
\noindent\textbf{Adversarial Learning}
Adversarial learning is an effective solution to improve the robustness of the deep neural network. In fully supervised models, adversarial learning can measure higher-order inconsistencies between labels and predictions \cite{kurakin2016adversarial}. Specifically, Li \etal \cite{li2019rosa} introduced generic noise to destroy adversarial perturbation of the input image for salient object detection. \cite{liu2020salient,tang2019salient} built the Generative Adversarial Network (GAN) \cite{gan_raw} based saliency detection network, where the fully connected discriminator was used to distinguish the real saliency map (ground truth) from the fake saliency map (prediction). Similarly, Jiang \etal \cite{jiang2020cmsalgan} designed a GAN based framework for RGB-D saliency detection to solve the cross-modality detection problem.
In the case of insufficient annotations, adversarial learning
can serve as guidance to select confident samples or generate new samples.
Souly \etal \cite{souly2017semi} used adversarial learning to generate fake images with image-level labels and noise, which in turn can make the real samples gathering in feature space and improve
accuracy for weakly supervised semantic segmentation.
\cite{hung2018adversarial,mittal2019semi} treated adversarial learning as a confidence measure to
obtain the trustworthy regions of semantic segmentation predictions of unlabeled data for semi-supervised learning.

Different from existing multi-task learning frameworks that mainly benefit from positively related tasks, we instead build the connections of two contradicting tasks within a joint learning framework by explicitly modeling the contradicting attributes with a similarity measure module.


\section{Our Method}
We design an uncertainty-aware joint learning framework as shown in Fig.~\ref{fig:network_overview} to learn SOD and COD in a unified framework. Firstly, as a data augmentation technique, we
select a group of easy samples from the COD training dataset
to achieve robust SOD.
Then, we present the \enquote{Similarity measure} module to explicitly model the \enquote{contradicting} attributes of the two tasks. Lastly, we introduce our uncertainty-aware adversarial training network to produce interpretable predictions during testing, and achieve higher-order similarity measure during training.

\subsection{Data interaction as data augmentation}
As shown in Fig.~\ref{fig:relationship of sod and cod}, there exist samples in the COD dataset that are both salient and camouflaged. We argue that those samples can be treated as hard samples for salient object detection to achieve robust learning. To select those samples from the COD dataset, we resort to Mean Absolute Error (MAE), and select samples in COD dataset \cite{fan2020camouflaged} which achieve the smallest MAE by testing it using a trained SOD model \cite{scrn_sal}. Specifically, for camouflaged object detection training dataset $D_c =\{X_i^c, Y_i^c\}_{i=1}^{N_c}$, where $i$ indexes the images, and $N_c$ is the size of camouflaged object detection training set. We defined the trained SOD model as $M_{\theta_s}$. Then we obtain saliency prediction of the images in $D_c$ as $P^c_s=M_{\theta_s}(X^c)=\{p^c_i\}_{i=1}^{N_c}$, where $p_i^c$ is the saliency prediction in COD training dataset. We assume that easy samples for COD can be treated as hard samples for SOD as shown in Fig.~\ref{fig:relationship of sod and cod}. Then we select $M=400$ samples $D_c^M$ of the smallest MAE in $D_c$, and replace it with $M=400$  randomly selected samples in our SOD training dataset \cite{wang2017learning} as a data augmentation technique. We show the selected samples in Fig.~\ref{fig:selected_easy_samples_from_cod}, which clearly illustrates the partially positive connection of the two tasks at the dataset level.

\begin{figure}[tp]
   \begin{center}
   \begin{tabular}{{c@{ } c@{ } c@{ } c@{ }}}
   {\includegraphics[height=0.16\linewidth]{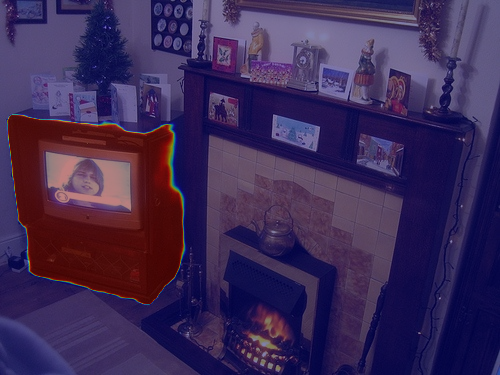}}&
   {\includegraphics[height=0.16\linewidth]{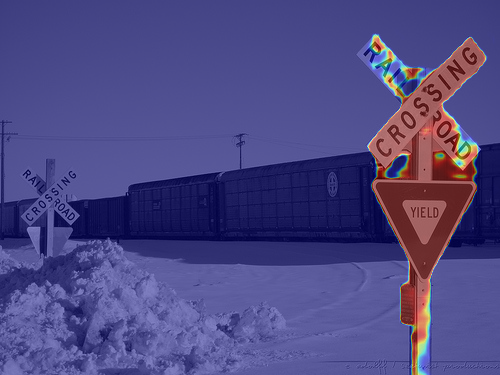}}&
   {\includegraphics[height=0.16\linewidth]{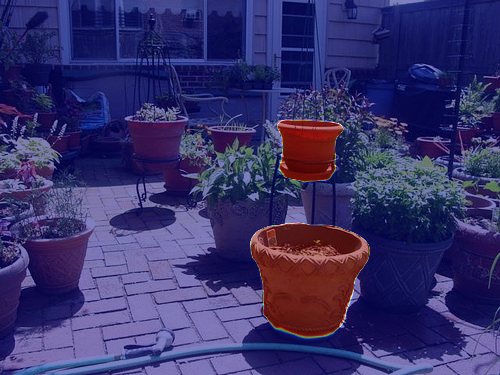}}&
   {\includegraphics[height=0.16\linewidth]{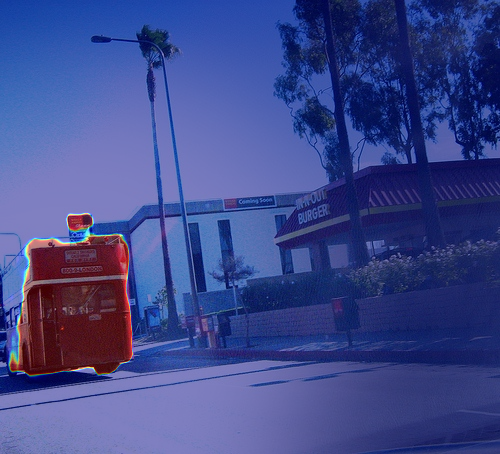}} \\
   {\includegraphics[height=0.16\linewidth]{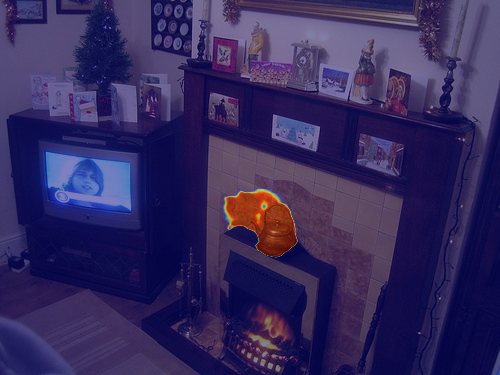}}&
   {\includegraphics[height=0.16\linewidth]{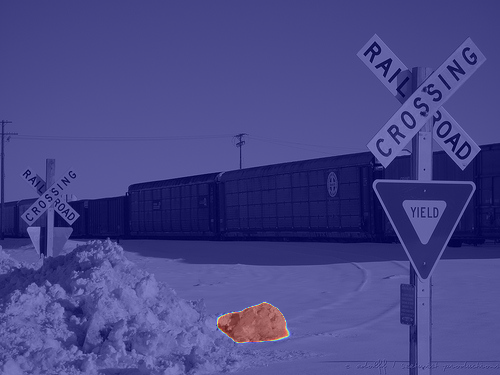}}&
   {\includegraphics[height=0.16\linewidth]{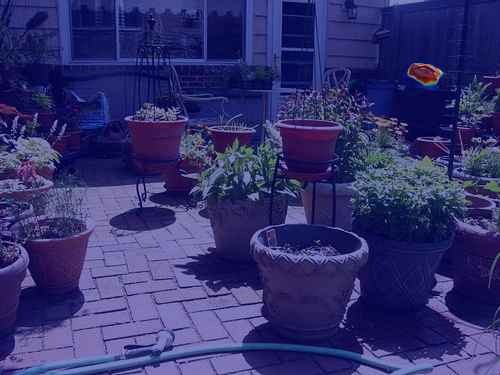}}&
   {\includegraphics[height=0.16\linewidth]{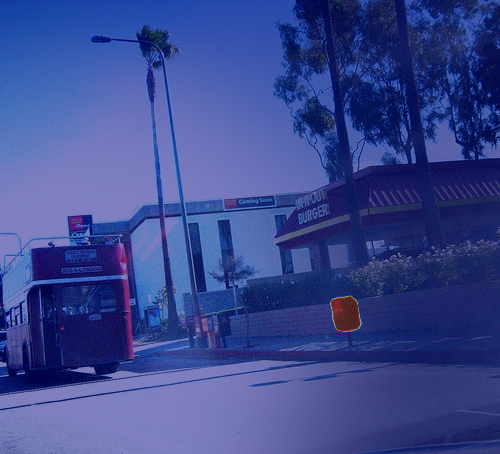}} \\
   
   \end{tabular}
   \end{center}
    \caption{Detected region from the saliency encoder branch (first row) and the camouflage encoder branch (second row).
    } 
    \label{fig:activation_maps_cod_sod}
\end{figure}

\subsection{Contradicting modeling} 
Similar as above, let's define our camouflaged object detection training dataset as $D_c=\{X_i^c,Y_i^c\}_{i=1}^{N_c}$ and augmented salient object detection training dataset as $D_s=\{X_i^s,Y_i^s\}_{i=1}^{N_s}$, where $i$ indexes images, $X$ and $Y$ are the image and ground truth pair, $N_c$ and $N_s$ are size of the camouflage training set and saliency training set respectively. Based on the \enquote{contradicting} attribute of SOD and COD, we design a \enquote{Similarity measure} module in Fig.~\ref{fig:network_overview} to explicitly model the connection of the two tasks.


Specifically, we introduce another set of images from PASCAL VOC 2007 dataset \cite{everingham2007pascal} as \enquote{connection modeling} dataset $D_p=\{X_i^p\}_{i=1}^{N_p}$, from which we extract the camouflaged feature and the salient feature. With the three datasets (COD dataset $D_c$, augmented SOD dataset $D_s$ and connection modeling dataset $D_p$), our contradicting modeling framework uses the \enquote{Feature encoder} module to extract both camouflage feature and saliency feature, and then use the \enquote{Similarity measure} module to model connection of the two tasks with the connection modeling dataset.

\noindent\textbf{Feature Encoder}\\
We design both the saliency encoder $E_{\alpha_s}$ and camouflage encoder $E_{\alpha_c}$ network with the same backbone network, \eg the ResNet50 \cite{he2016deep}, where $\alpha_s$ and $\alpha_c$ are network parameter sets of each of them respectively. Initially, the ResNet50 backbone network has four groups\footnote{We define feature maps of the same spatial size as
same group.} of convolutional layers of channel size 256, 512, 1024 and 2048 respectively.
We then define the output features of both encoders as $F_{\alpha_s}=\{f^s_1,f^s_2,f^s_3,f^s_4\}$  and $F_{\alpha_c}=\{f^c_1,f^c_2,f^c_3,f^c_4\}$, where $f_{k}, k=1,...,4$ is feature map of the $k$-th group.

\noindent\textbf{Similarity Measure}\\
Different from the feature encoder module, which takes images from $D_c$ and $D_s$ as input to produce task-specific feature maps, the similarity measure module $S_\theta$ takes the connection modeling data $D_p$ as input to model the connection of SOD and COD, where $\theta$ is parameter set of the similarity measure module.
Given the trained saliency encoder $E_{\alpha_s}$ and camouflage encoder $E_{\alpha_c}$, the saliency feature and camouflage feature of images $X^p$
are $F^p_{\alpha_s}=\{f^{p}_{s1},f^{p}_{s2},f^{p}_{s3},f^{p}_{s4}\}$ and $F^p_{\alpha_c}=\{f^{p}_{c1},f^{p}_{c2},f^{p}_{c3},f^{p}_{c4}\}$ respectively. We then concatenate each of above two features channel-wise and feed them to the same fully connected layer to obtain the latent saliency feature and latent camouflage feature of $X^p$ as $f^{sp}=S_\theta(F^p_{\alpha_s})$ and $f^{cp}=S_\theta(F^p_{\alpha_c})$ respectively. Empirically, we set the dimension of the latent space as $K=700$.
For the same image in $D_p$, we assume that the SOD network and COD network should focus on different regions, leading to different feature representation.
Then, we choose the cosine similarity to measure the difference between the saliency feature and the camouflage feature in latent space, and define the 
latent space loss as:
\begin{equation}
\label{latent_loss}
   \mathcal{L}_{latent}=cos(f^{sp},f^{cp})=\frac{f^{sp}\cdot f^{cp}}{\left \|  f^{sp}\right \|\times\left \| f^{cp}\right \| }.
\end{equation}

In Fig.~\ref{fig:activation_maps_cod_sod}, we show the activation region (the processed predictions) of the same image from both the saliency encoder (first row) and camouflage encoder (second row). Specifically, given same image $X^p$, we compute it's camouflage map and saliency map, and highlight the detected foreground region in red. 
Fig.~\ref{fig:activation_maps_cod_sod}  shows that the two encoders focus on different regions of the image, where the saliency encoder pays more attention to the region that stand out from the context,
and the camouflage encoder focuses more on the hidden object with similar color or structure as the background,
which is consistent with our assumption that these two tasks are contradicting with each other in general.


\subsection{Uncertainty-aware adversarial learning}
\begin{figure}[tp]
   \begin{center}
   \begin{tabular}{{c@{ } c@{ } c@{ } c@{ }}}
   {\includegraphics[height=0.21\linewidth]{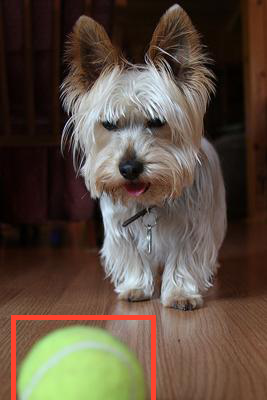}}&
    {\includegraphics[height=0.21\linewidth]{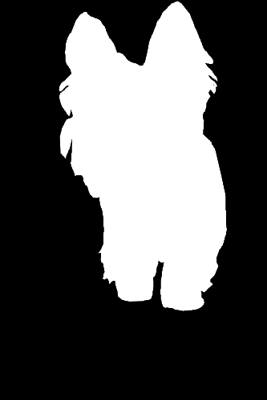}}&
    {\includegraphics[height=0.21\linewidth]{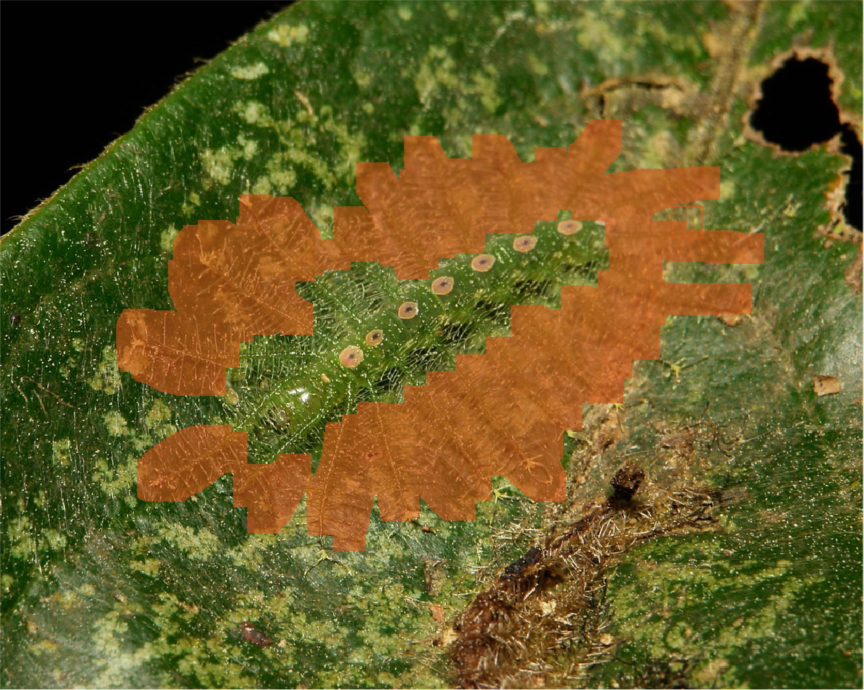}}&
    {\includegraphics[height=0.21\linewidth]{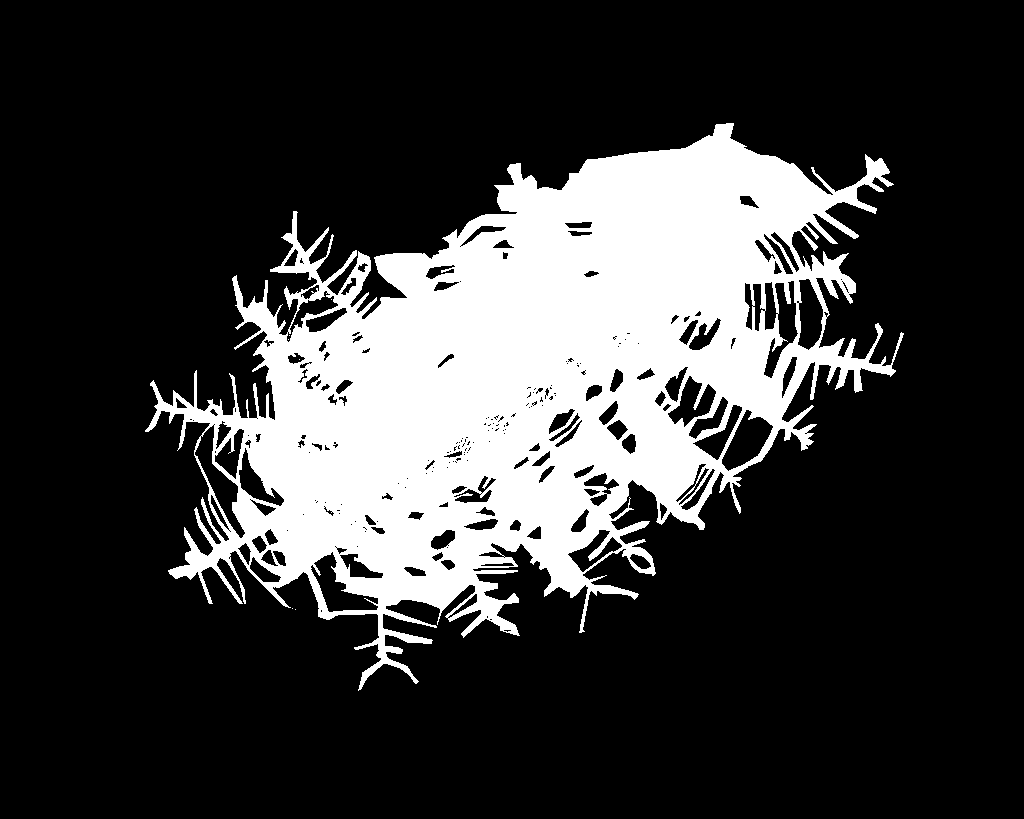}} \\
    \footnotesize{(a)} &
     \footnotesize{(b)} &\footnotesize{(c)} &\footnotesize{(d)} \\
   \end{tabular}
   \end{center}
    \caption{Uncertainty in labeling the SOD dataset ((a) and (b)) and the COD dataset ((c),(d))} 
    \label{fig:labeling_uncertainty}
\end{figure}

As shown in Fig.~\ref{fig:labeling_uncertainty}, for the SOD dataset, the uncertainty comes from the ambiguity of saliency ((a) (b)), and for the COD dataset, the uncertainty results from the difficulty of annotation ((c) (d)). \eg the ball in the orange rectangle (a) can be defined as salient, but it's background in (b). The orange region in (c) belongs to the camouflaged object, while it's too similar to the background, making it very difficult to create the accurate annotation. We then introduce an uncertainty-aware adversarial training strategy to model the task-specific uncertainty in our joint learning framework, which includes a \enquote{Prediction decoder} module to produce task-related predictions, a \enquote{Confidence estimation} module to estimate uncertainty of each prediction, and an adversarial learning strategy for robust model training.

\noindent\textbf{Prediction decoder}\\
As shown in Fig.~\ref{fig:network_overview}, we design a shared decoder structure for joint SOD and COD learning. We argue that the different \enquote{Feature encoder} modules can generate task-specific features for COD images and SOD images. Then the \enquote{Prediction decoder} module aims to integrate the task-specific feature with their corresponding lower level feature to produce predictions. 
Specifically, given task-specific feature $F_{\alpha_s}=\{f^s_1,f^s_2,f^s_3,f^s_4\}$ and $F_{\alpha_c}=\{f^c_1,f^c_2,f^c_3,f^c_4\}$ from the saliency encoder and camouflage encoder respectively, the prediction decoder produces saliency map $G_{\beta}(F_{\alpha_s})$ and camouflage map $G_{\beta}(F_{\alpha_c})$, where $\beta$ is parameter set of the prediction decoder module. Specifically, we design a top-down connection network with the residual channel attention module \cite{zhang2018image} $Re$ to extract finer features. Furthermore, the dual attention module \cite{dual_attention} $Da$ is adopted to effectively fuse higher level semantic information with the lower level structure information to obtain the initial predictions:
\begin{equation}
    \label{init_decoder_sod}
    \begin{aligned}
       G_{init}(F_{\alpha_s}) =C_{cla}(Re[Da(f^s_4), f^s_{43\_init},Da(f^s_2)]),
    \end{aligned}
\end{equation}\begin{equation}
    \label{init_decoder_cod}
    \begin{aligned}
       G_{init}(F_{\alpha_c}) = C_{cla}(Re[Da(f^c_4),f^c_{43\_init},Da(f^c_2)]),
    \end{aligned}
\end{equation}
where $f_{43\_init}=conv(Re[Da(f_4),Da(f_3)])$, $conv$ is the $3\times3$ convolutional layer of output channel size $C=32$, $[.]$ is the channel-wise concatenation operation, and we upsample the features to the same spatial size before concatenation. $C_{cla}$ is the classification layer of kernel size $3\times3$, which maps the feature map to one channel prediction for each task. Then we add a refined structure to the decoder network in order to obtain a detailed prediction $G_{\beta}(F_{\alpha_s})$ and $G_{\beta}(F_{\alpha_c})$, we use the holistic attention module  \cite{cpd_sal} $Ha$ to integrate features:
\begin{equation}
    \begin{aligned}
       f_{r2}=Ha(G_{init}(F_{\alpha}),f_2),f_{r3}=R_3(f_{r2}), f_{r4}=R_4(f_{r3}),\\
    \end{aligned}
\end{equation}
where $R_3$ and $R_4$ is the ResNet50 backbone convolutional layers of channel size 1024 and 2048 respectively.
Then, we obtain the task-specific predictions $G_{\beta}(F_{\alpha_s})$ and $G_{\beta}(F_{\alpha_c})$:
\begin{equation}
    \label{refine_decoder_sod}
    \begin{aligned}
    G_{\beta}(F_{\alpha_s}) = C_{cla}(Re[Da(f^s_{r4}),f^s_{43},f^s_{432},conv(f^s_1)]),
    \end{aligned}
\end{equation}
\begin{equation}
    \label{refine_decoder_cod}
    \begin{aligned}
    G_{\beta}(F_{\alpha_c}) = C_{cla}(Re[Da(f^c_{r4}),f^c_{43},f^c_{432},conv(f^c_1)]),
    \end{aligned}
\end{equation}
where the feature $f_{43}=conv(Re[Da(f_{r4}),Da(f_{r3}])$, and the feature $f_{432}=conv(Re[Da(f_{r4}),f_{43},Da(f_{r2})])$.

\noindent\textbf{Confidence Estimation}\\
As discussed above, uncertainty exists in both SOD and COD dataset. We introduce the \enquote{Confidence estimation} module to explicitly model the confidence of network predictions.
Specifically, we design a fully convolutional discriminator network to evaluate confidence of the predictions from the \enquote{Prediction decoder} module.
The fully convolutional discriminator network $D^f_\gamma$ consists of five convolution layers as shown in Table~\ref{tab:discriminator_gan}, and produce a one-channel confidence map, where $\gamma$ is the network parameter set. Note that, we have the batch normalization and leaky relu layers after the first four convolutional layers. $D^f_\gamma$ aims to produce all-zero output with prediction $G_{\beta}(F_{\alpha_s})$ or $G_{\beta}(F_{\alpha_c})$ as input, and all-one matrix with ground truth as input.

\noindent\textbf{Adversarial Learning}\\
We introduce adversarial learning to learn both the \enquote{Prediction decoder} and the \enquote{Confidence estimation} modules.

\begin{table}[!htp]
  \centering
  \footnotesize
  \renewcommand{\arraystretch}{1.1}
  \renewcommand{\tabcolsep}{1.5mm}
  \caption{Network structure of the discriminator network.}
  \begin{tabular}{c|c|c|c|c}
  \hline
  \multicolumn{1}{c|}{Input Channel}&\multicolumn{1}{c|}{Output Channel}&\multicolumn{1}{c|}{kernel size}&\multicolumn{1}{c|}{Stride}&\multicolumn{1}{c}{Padding} \\
  \hline
  1 & 64 & 3 & 2 & 1 \\ 
  64 & 64 & 3 & 1 & 1 \\
  64 & 64 & 3 & 2 & 1 \\
  64 & 64 & 3 & 1 & 1 \\
  64 & 1 & 3 & 2 & 1 \\
  \hline
  \end{tabular}
  \label{tab:discriminator_gan}
\end{table}



For the prediction decoder module, we first have the task-specific loss function to learn each task. Specifically, we adopt the structure-aware loss function \cite{wei2020f3net} for both SOD and COD, and define the loss function as:
\begin{equation}
\label{structure_loss}
   \mathcal{L}_{str}(Pred,Y)=\omega*\mathcal{L}_{ce}(Pred,Y)+\mathcal{L}_{iou}(Pred,Y), 
\end{equation}
where $\omega$ is the edge-aware weight, which is defined as $\omega=1+5*\left | (avg\_pool(Y)-Y)\right |$, $\mathcal{L}_{ce}$ is the cross-entropy loss,  $\mathcal{L}_{iou}$ is the boundary-IOU loss \cite{nldf_sal}, which is defined as:
\begin{equation}
   \mathcal{L}_{iou}= 1-\frac{\omega*inter+1}{\omega*union - \omega*inter+1},
\end{equation}
where $inter=Pred*Y$, and $union=Pred+Y$.

And we have the structure-aware loss function for SOD and COD as:
\begin{equation}
\label{structure_loss_sod}
  \mathcal{L}_{str}^s=0.5*[{L}_{str}(G_{init}(F_{\alpha_s}),Y^s)+{L}_{str}(G_{\beta}(F_{\alpha_s}),Y^s)], 
\end{equation}
\begin{equation}
\label{structure_loss_cod}
   \mathcal{L}_{str}^c=0.5*[{L}_{str}(G_{init}(F_{\alpha_c}),Y^c)+{L}_{str}(G_{\beta}(F_{\alpha_c}),Y^c)]. 
\end{equation}

To achieve adversarial learning, we have adversarial loss for both SOD and COD, which is defined as cross-entropy loss between network prediction and the pre-defined \enquote{real} indicator as:
\begin{equation}
    \label{adv_pred_sod}
    \mathcal{L}_{adv}^s = \mathcal{L}_{ce}(D_\gamma^f(G_\beta(F_{\alpha_s})), \mathbf{1}),
\end{equation}
\begin{equation}
    \label{adv_pred_cod}
    \mathcal{L}_{adv}^c = \mathcal{L}_{ce}(D_\gamma^f(G_\beta(F_{\alpha_c})), \mathbf{1}),
\end{equation}
respectvely for each task, where $\mathbf{1}$ is an all-one matrix. In this way, the discriminator takes model prediction as input, and tries to recognize it as real ground truth.

For the confidence estimation module, similar to the typical definition of discriminator in GAN \cite{gan_raw}, we want it to clearly distinguish model prediction and ground truth map. Then, the adversarial loss for the confidence estimation module of the SOD task is defined as:
\begin{equation}
\label{dis_sod}
    \mathcal{L}_{dis}^s=\mathcal{L}_{ce}(D_\gamma^f(G_\beta(F_{\alpha_s})),\mathbf{0}) + \mathcal{L}_{ce}(D_\gamma^f(Y^s),\mathbf{1}).
\end{equation}
We then have the adversarial loss for the COD task as:
\begin{equation}
\label{dis_cod}
    \mathcal{L}_{dis}^c=\mathcal{L}_{ce}(D_\gamma^f(G_\beta(F_{\alpha_c})),\mathbf{0}) + \mathcal{L}_{ce}(D_\gamma^f(Y^c),\mathbf{1}).
\end{equation}


\noindent\textbf{Objective Function}\\
\begin{algorithm}[!htp]
\small
\caption{Uncertainty-aware joint salient object detection and camouflaged object detection}
\textbf{Input}: Training image sets: $D_s = \{X^s,Y^s\}$, $D_c = \{X^c,Y^c\}$ and $D_p = \{X^p\}$; Maximal number of learning iterations $T$. \\
\textbf{Output}: 
$\alpha_s$, $\alpha_c$ for feature encoder, $\theta$ for similarity measure, $\beta$ for prediction decoder, and $\gamma$ for confidence estimation;
\begin{algorithmic}[1]
\State Initialize $\alpha_s$, $\alpha_c$ with ResNet50 \cite{he2016deep}, and $\theta$, $\beta$, $\gamma$ randomly.
\For{$t \leftarrow  1$ to $T$}
\State Compute latent loss in Eq.~\eqref{latent_loss} and update $\alpha_s$, $\alpha_c$, $\theta$
\State Compute generator loss and adversarial loss for SOD according to Eq.~\eqref{sal_update}, and update $\alpha_s$, $\beta$
\State Compute adversarial loss for confidence estimation module according to Eq.~\eqref{dis_sod} and update $\gamma$
\State Compute generator loss and adversarial loss for COD according to Eq.~\eqref{cod_update}, and update $\alpha_c$, $\beta$
\State Compute adversarial loss for confidence estimation module according to Eq.~\eqref{dis_cod} and update $\gamma$
\EndFor
\end{algorithmic} 
\label{our_alg}
\end{algorithm}
As shown in Fig.~\ref{fig:network_overview}, the two tasks have separate encoder, shared decoder as well as the shared confidence estimation module. Given camouflaged object detection training images in $D_c$, and salient object detection training images in $D_s$, as well as the connection modeling images in $D_p$, we first compute the similarity measure of $D_p$ as Eq.~\eqref{latent_loss} and update the feature encoder ($\alpha_c$ and $\alpha_s$) and similarity measure module ($\theta$). Then, we train the adversarial learning for the saliency generator branch (saliency encoder and prediction decoder)
with loss function:
\begin{equation}
    \label{sal_update}
    \mathcal{L}_{sod} = \mathcal{L}_{str}^s + \lambda_1 \mathcal{L}_{adv}^s,
\end{equation}
where $\lambda_1$ is a trade-off parameter, and empirically we set $\lambda_1=0.01$ as \cite{hung2018adversarial}.
Similarly, we train the adversarial learning for the camouflage generator branch (camouflage encoder and prediction decoder)
with loss function:
\begin{equation}
    \label{cod_update}
    \mathcal{L}_{cod} = \mathcal{L}_{str}^c + \lambda_2 \mathcal{L}_{adv}^c,
\end{equation}
where we set $\lambda_2=0.01$ in our experiments.

Then, we train the confidence estimation module of parameter set $\gamma$ with the loss function:
\begin{equation}
    \label{confidence_estimation_update}
    \mathcal{L}_{conf} =  \mathcal{L}_{dis}^s +  \mathcal{L}_{dis}^c.
\end{equation}
Algorithm \ref{our_alg} is our complete training algorithm.

\begin{table*}[t!]
  \centering
  \scriptsize
  \renewcommand{\arraystretch}{1.05}
  \renewcommand{\tabcolsep}{0.3mm}
  \caption{Performance comparison with benchmark saliency detection models.
  }
  \begin{tabular}{l|cccc|cccc|cccc|cccc|cccc|cccc}
  \hline
  &\multicolumn{4}{c|}{DUTS}&\multicolumn{4}{c|}{ECSSD}&\multicolumn{4}{c|}{DUT}&\multicolumn{4}{c|}{HKU-IS}&\multicolumn{4}{c|}{THUR}&\multicolumn{4}{c}{SOC} \\
    Method & $S_{\alpha}\uparrow$&$F_{\beta}\uparrow$&$E_{\xi}\uparrow$&$\mathcal{M}\downarrow$& $S_{\alpha}\uparrow$&$F_{\beta}\uparrow$&$E_{\xi}\uparrow$&$\mathcal{M}\downarrow$& $S_{\alpha}\uparrow$&$F_{\beta}\uparrow$&$E_{\xi}\uparrow$&$\mathcal{M}\downarrow$& $S_{\alpha}\uparrow$&$F_{\beta}\uparrow$&$E_{\xi}\uparrow$&$\mathcal{M}\downarrow$& $S_{\alpha}\uparrow$&$F_{\beta}\uparrow$&$E_{\xi}\uparrow$&$\mathcal{M}\downarrow$& $S_{\alpha}\uparrow$&$F_{\beta}\uparrow$&$E_{\xi}\uparrow$&$\mathcal{M}\downarrow$ \\ \hline
    NLDF \cite{nldf_sal} & .816 & .757 & .851 & .065 & .870 & .871 & .896 & .066 & .770 & .683 & .798 & .080 & .879 & .871 & .914 & .048 & .801 & .711 & .827 & .081  & .816 & .319 & .837 & .106\\ 
   PiCANet \cite{picanet} & .842 & .757 & .853 & .062 & .898 & .872 & .909 & .054 & .817 & .711 & .823 & .072 & .895 & .854 & .910 & .046 & .818 & .710 & .821 & .084 & .801 & .332 & .810 & .133 \\ 
   CPD \cite{cpd_sal} & .869 & .821 & .898 & .043 & .913 & .909 & .937 & .040 & .825 & .742 & .847 & .056 & .906 & .892 & .938 & .034 & .835 & .750 & .853 & .068 & .841 & .356 & .862 & .093  \\
   SCRN \cite{scrn_sal} & .885 & .833 & .900 & .040 & .920 & .910 & .933 & .041 & .837 & .749 & .847 & .056 & .916 & .894 & .935 & .034 & .845 & .758 & .858 & .066  & .838 & .363 & .859 & .099\\ 
   PoolNet \cite{Liu19PoolNet} & .887 & .840 & .910 & .037 & .919 & .913 & .938 & .038 & .831 & .748 & .848 & .054 & .919 & .903 & .945 & .030 & .834 & .745 & .850 & .070  & .829 & .355 & .846 & .106 \\ 
    BASNet \cite{basnet_sal} & .876 & .823 & .896 & .048 & .910 & .913 & .938 & .040 & .836 & .767 & .865 & .057 & .909 & .903 & .943 & .032 & .823 & .737 & .841 & .073  & .841 & .359 & .864 & .092\\ 
   EGNet \cite{zhao2019EGNet} & .878 & .824 & .898 & .043 & .914 & .906 & .933 & .043 & .840 & .755 & .855 & .054 & .917 & .900 & .943 & .031 & .839 & .752 & .854 & .068 & \textbf{.858} & .353 & \textbf{.873} & \textbf{.078}  \\   
     AFNet \cite{feng2019attentive} & .867 & .812 & .893 & .046 & .907 & .901 & .929 & .045 & .826 & .743 & .846 & .057 & .905 & .888 & .934 & .036 & .825 & .733 & .840 & .072  & .700 & .062 & .684 & .115\\

      CSNet \cite{gao2020highly} & .884 & .834 & .907 & .040 & .920 & .911 & .940 & .038 & .836 & .750 & .852 & .055 & .918 & .900 & .944 & .031 & .841 & .756 & .856 & .068 & .834 & .352 & .850 & .103  \\
   F3Net \cite{wei2020f3net} & .888 & .852 & .920 & .035 & .919 & .921 & .943 & .036 & .839 &  .766 & .864 & .053 & .917 & .910 & .952 & .028 & .838 & .761 & .858 & .066  & .828 & .340 & .846 & .098\\
   ITSD \cite{zhou2020interactive} & .886 & .841 & .917 & .039 & .920 & .916 & .943 & .037 & .842 & .767 & .867 & .056 & .921 & .906 & .950 & .030 & .836 & .753 & .852 & .070  & .773 & .361 & .792 & .166\\ 



   \hline
Ours & \textbf{.899} & \textbf{.866} & \textbf{.937} & \textbf{.032} & \textbf{.933} & \textbf{.935} & \textbf{.960} & \textbf{.030} & \textbf{.850} & \textbf{.782} & \textbf{.884} & \textbf{.051} & \textbf{.931} & \textbf{.924} & \textbf{.867} & \textbf{.026} & \textbf{.849} & \textbf{.774} & \textbf{.872} & \textbf{.065} & .845 & .\textbf{374} & .856 & .092  \\

   \hline 
  \end{tabular}
  \label{tab:benchmark_sod_model_comparison}
\end{table*}

\begin{figure*}[tp]
   \begin{center}
   \begin{tabular}{{c@{ } c@{ } c@{ } c@{ } c@{ } c@{ } c@{ } c@{ } c@{ }}}
    {\includegraphics[width=0.100\linewidth]{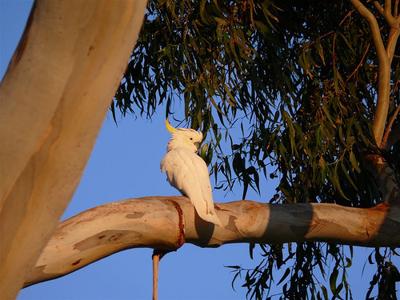}}&
    {\includegraphics[width=0.100\linewidth]{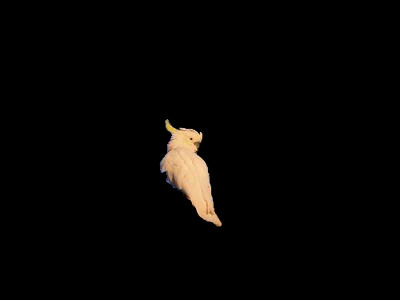}}&
    {\includegraphics[width=0.100\linewidth]{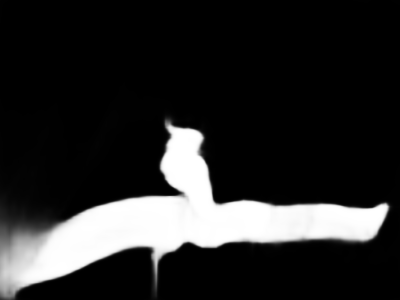}}&
   {\includegraphics[width=0.100\linewidth]{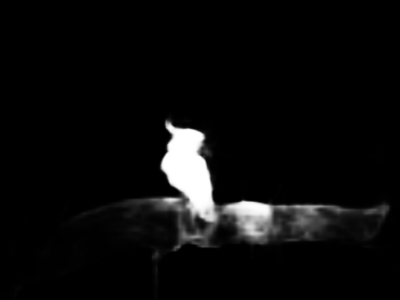}}&
    {\includegraphics[width=0.100\linewidth]{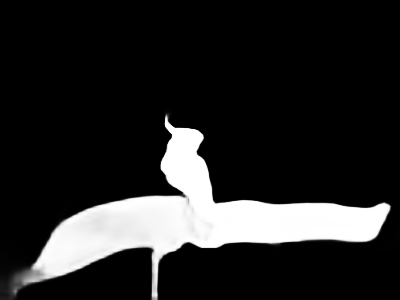}}&
    {\includegraphics[width=0.100\linewidth]{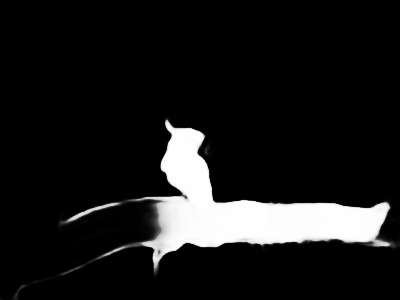}}&
    {\includegraphics[width=0.100\linewidth]{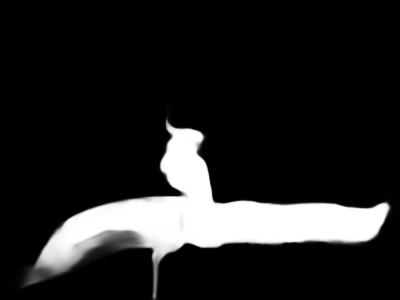}}&
    {\includegraphics[width=0.100\linewidth]{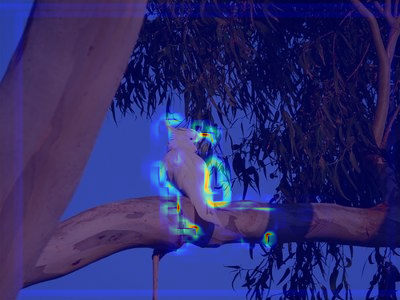}}&
    {\includegraphics[width=0.100\linewidth]{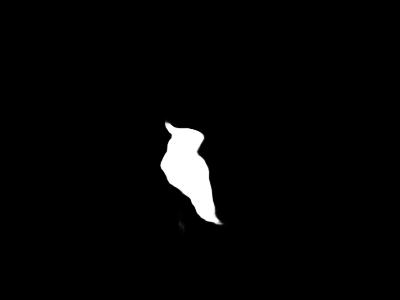}}\\
    {\includegraphics[width=0.100\linewidth]{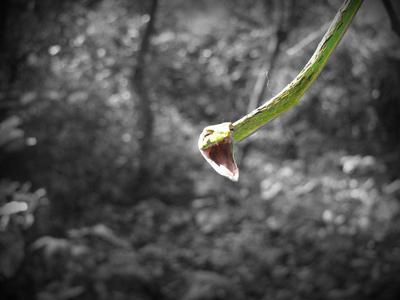}}&
    {\includegraphics[width=0.100\linewidth]{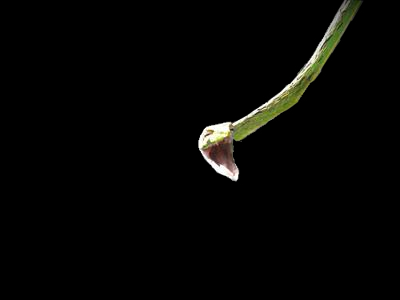}}&
    {\includegraphics[width=0.100\linewidth]{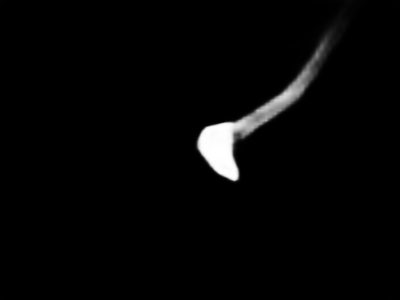}}&
   {\includegraphics[width=0.100\linewidth]{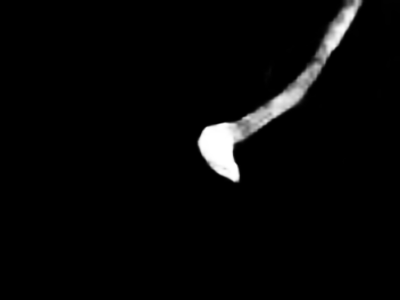}}&
    {\includegraphics[width=0.100\linewidth]{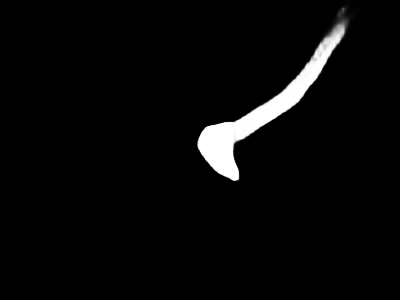}}&
    {\includegraphics[width=0.100\linewidth]{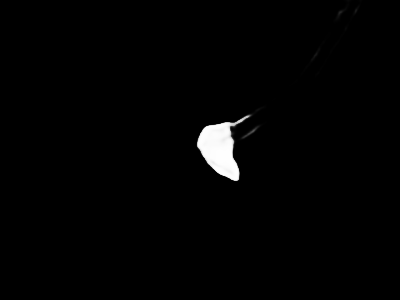}}&
    {\includegraphics[width=0.100\linewidth]{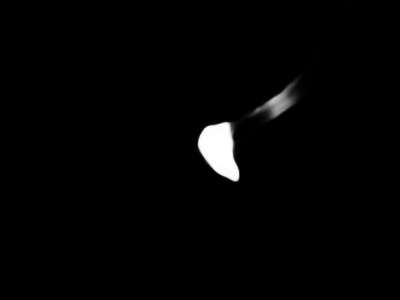}}&
    {\includegraphics[width=0.100\linewidth]{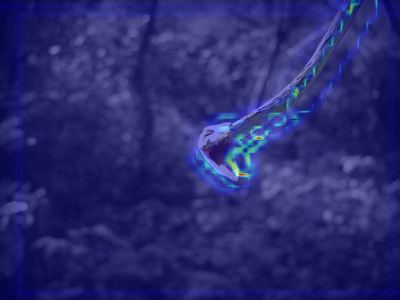}}&
    {\includegraphics[width=0.100\linewidth]{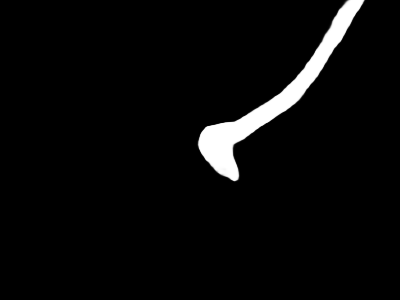}}\\
    \footnotesize{Image} &
     \footnotesize{GT} &\footnotesize{SCRN \cite{scrn_sal}} &\footnotesize{CSNet \cite{gao2020highly}}&\footnotesize{F3Net \cite{wei2020f3net}}&\footnotesize{ITSD \cite{zhou2020interactive}} &\footnotesize{EGNet \cite{zhao2019EGNet}} &\footnotesize{Uncertainty} &\footnotesize{Ours}\\
   \end{tabular}
   \end{center}
    \caption{Predictions of competing salient object detection models and ours.
    } 
    \label{fig:sod_samples_comparison}
\end{figure*}

\begin{table*}[t!]
  \centering
  \scriptsize
  \renewcommand{\arraystretch}{1.10}
  \renewcommand{\tabcolsep}{2.3mm}
  \caption{Performance comparison with re-implemented camouflaged object detection models.}
  \begin{tabular}{l|cccc|cccc|cccc}
  \hline
  &\multicolumn{4}{c|}{CAMO}&\multicolumn{4}{c|}{CHAMELEON}&\multicolumn{4}{c}{COD10K} \\
    Method & $S_{\alpha}\uparrow$&$F_{\beta}\uparrow$&$E_{\xi}\uparrow$&$\mathcal{M}\downarrow$& $S_{\alpha}\uparrow$&$F_{\beta}\uparrow$&$E_{\xi}\uparrow$&$\mathcal{M}\downarrow$ &  $S_{\alpha}\uparrow$ & $F_{\beta}\uparrow$ & $E_{\xi}\uparrow$ & $\mathcal{M}\downarrow$  \\
  \hline
  NLDF\cite{nldf_sal}& 0.665 & 0.564 & 0.664 & 0.123 & 0.798 & 0.714 & 0.809 & 0.063 & 0.701 & 0.539 & 0.709 & 0.059   \\
  PiCANet\cite{picanet} & 0.701 & 0.573 & 0.716 & 0.125 & 0.765 & 0.618 & 0.779 & 0.085 & 0.696 & 0.489 & 0.712 & 0.081 \\ 
  CPD \cite{cpd_sal} & 0.716 & 0.618 & 0.723 & 0.113 & 0.857 & 0.771 & 0.874 & 0.048 & 0.750 & 0.595 & 0.776 & 0.053   \\
 SCRN \cite{scrn_sal}& 0.779 & 0.705 & 0.796 & 0.090 & 0.876 & 0.787 & 0.889 & 0.042 & 0.789 & 0.651  & 0.817 & 0.047   \\
   PoolNet \cite{Liu19PoolNet}  & 0.730 & 0.643 & 0.746 & 0.105 & 0.845 & 0.749 & 0.864 & 0.054 & 0.740 & 0.576 & 0.776 & 0.056   \\ 
   BASNet \cite{basnet_sal} & 0.615 & 0.503 & 0.671 & 0.124 & 0.847 & 0.795 & 0.883 & 0.044 & 0.661 & 0.486 & 0.729 & 0.071    \\

  EGNet \cite{zhao2019EGNet} & 0.737 & 0.655 & 0.758 & 0.102 & 0.856 & 0.766 & 0.883 & 0.049 & 0.751 & 0.595 & 0.793 & 0.053    \\

  CSNet\cite{gao2020highly} & 0.771 & 0.705 & 0.795 & 0.092 & 0.856 & 0.766 & 0.869 & 0.047 & 0.778 & 0.635 & 0.810 & 0.047    \\ 

  F3Net \cite{wei2020f3net} & 0.711 & 0.616 & 0.741 & 0.109 & 0.848 & 0.770 & 0.894 & 0.047 & 0.739 & 0.593 & 0.795 & 0.051  \\
  ITSD\cite{zhou2020interactive} & 0.750 & 0.663 & 0.779 & 0.102 & 0.814 & 0.705 & 0.844 & 0.057 & 0.767 & 0.615 & 0.808 & 0.051    \\ 
  SINet \cite{fan2020camouflaged} & 0.745 & 0.702 & 0.804 & 0.092 & 0.872 & 0.827 & 0.936 & 0.034 & 0.776 & 0.679 & 0.864 & 0.043 \\ \hline
  Ours  & \textbf{0.803} & \textbf{0.759} & \textbf{0.853} & \textbf{0.076} & \textbf{0.894} & \textbf{0.848} & \textbf{0.943} & \textbf{0.030} & \textbf{0.817} & \textbf{0.726} & \textbf{0.892} & \textbf{0.035}        \\ 
   \hline

  \end{tabular}
  \label{tab:benchmark_cod_model_comparison}
  \vspace{-2mm}
\end{table*}

\begin{figure*}[tp]
   \begin{center}
   \begin{tabular}{{c@{ } c@{ } c@{ } c@{ } c@{ } c@{ } c@{ } c@{ } c@{ }}}
   {\includegraphics[width=0.100\linewidth]{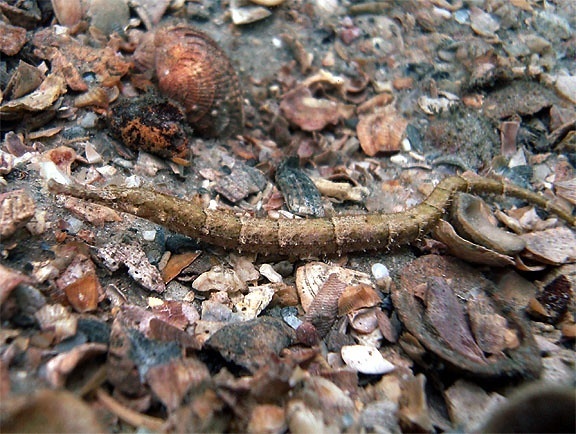}}&
    {\includegraphics[width=0.100\linewidth]{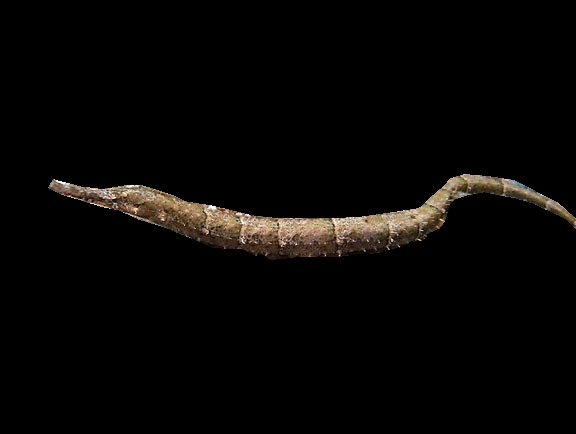}}&
     {\includegraphics[width=0.100\linewidth]{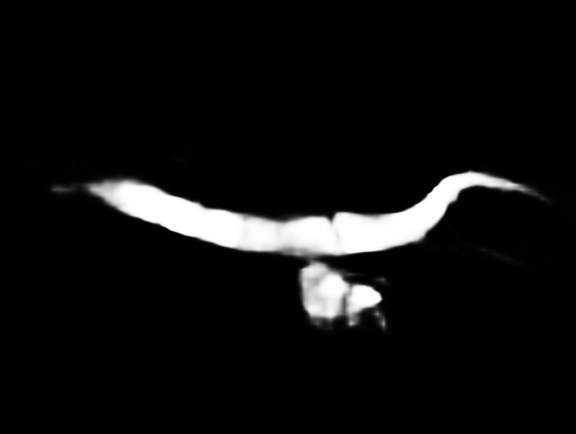}}&
    {\includegraphics[width=0.100\linewidth]{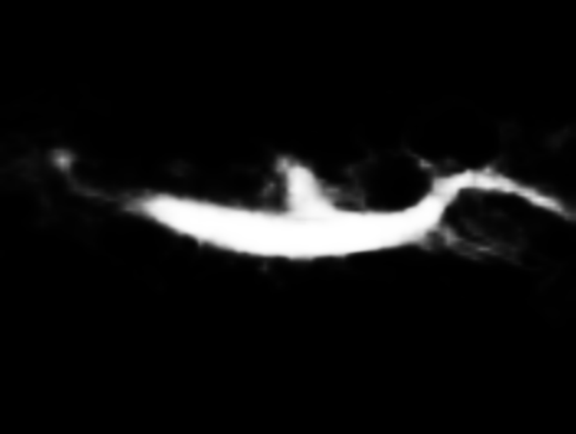}}&
    {\includegraphics[width=0.100\linewidth]{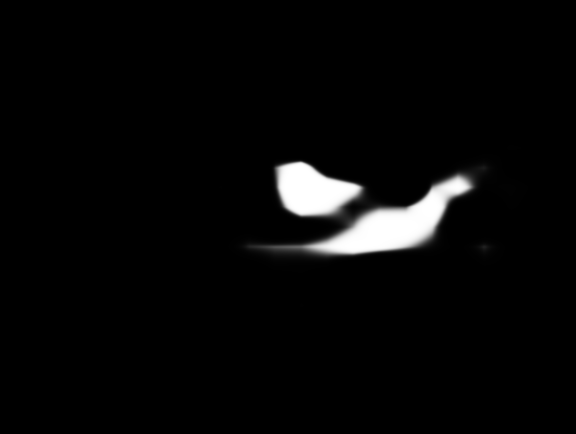}}&
    {\includegraphics[width=0.100\linewidth]{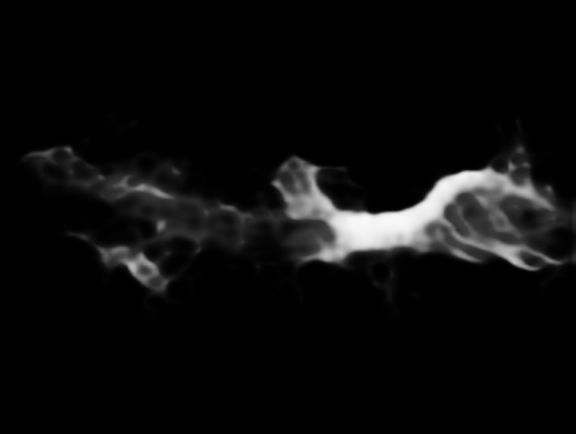}}&
    {\includegraphics[width=0.100\linewidth]{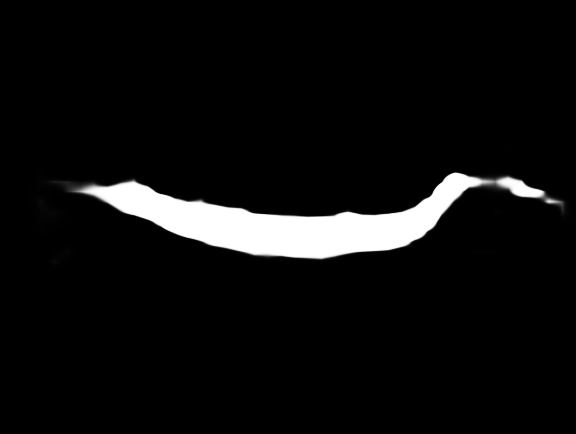}}&
    {\includegraphics[width=0.100\linewidth]{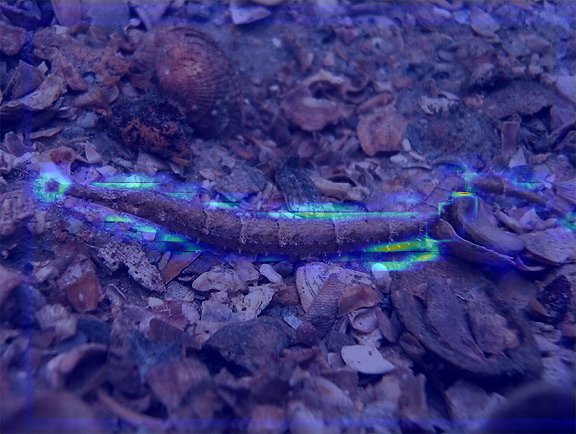}}&
    {\includegraphics[width=0.100\linewidth]{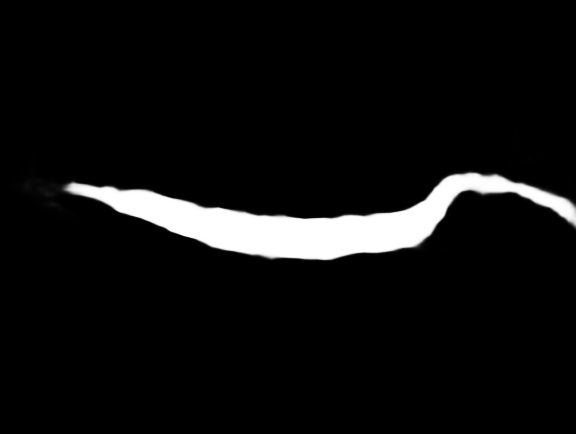}}\\
    {\includegraphics[width=0.100\linewidth]{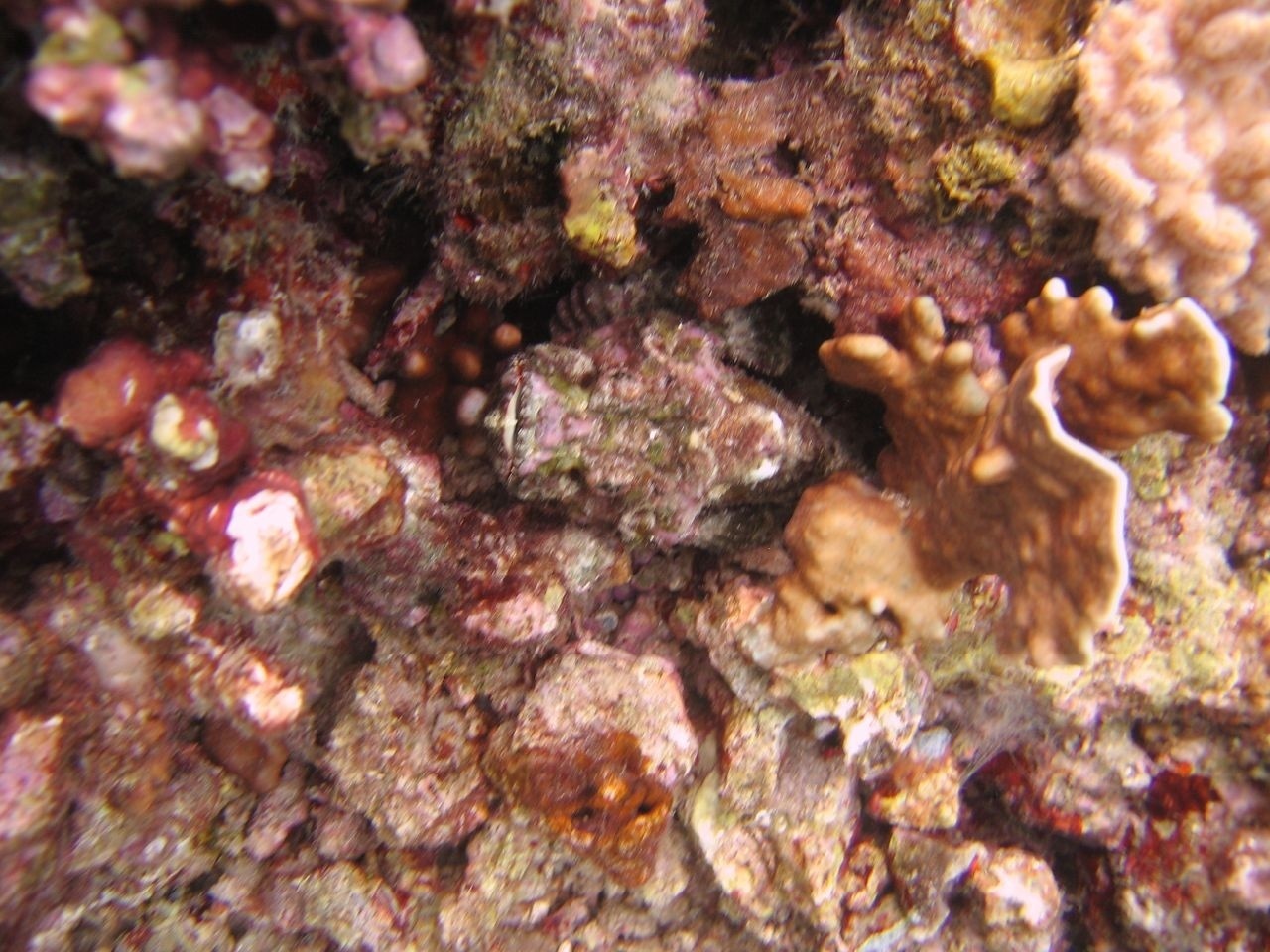}}&
     {\includegraphics[width=0.100\linewidth]{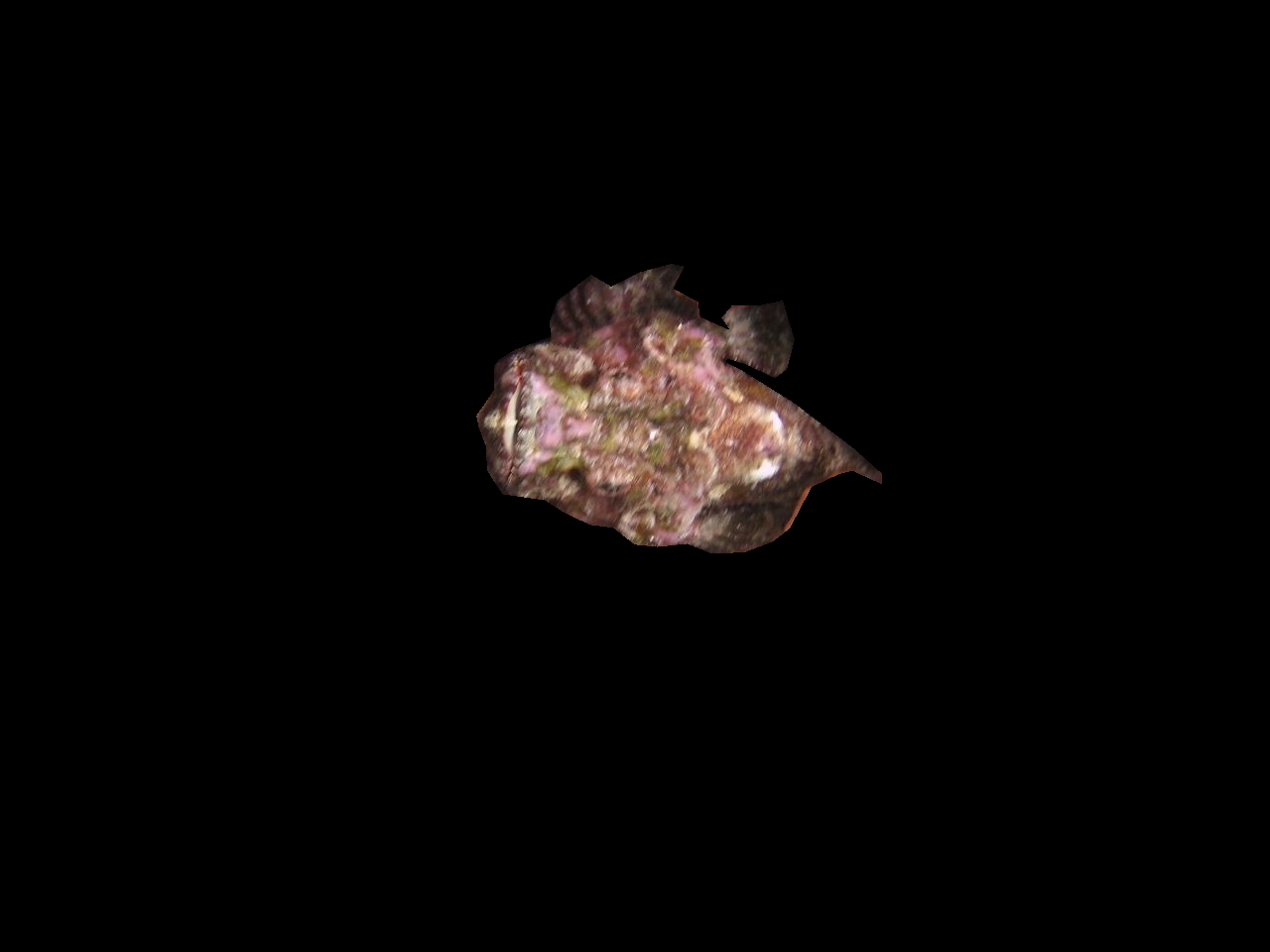}}&
    {\includegraphics[width=0.100\linewidth]{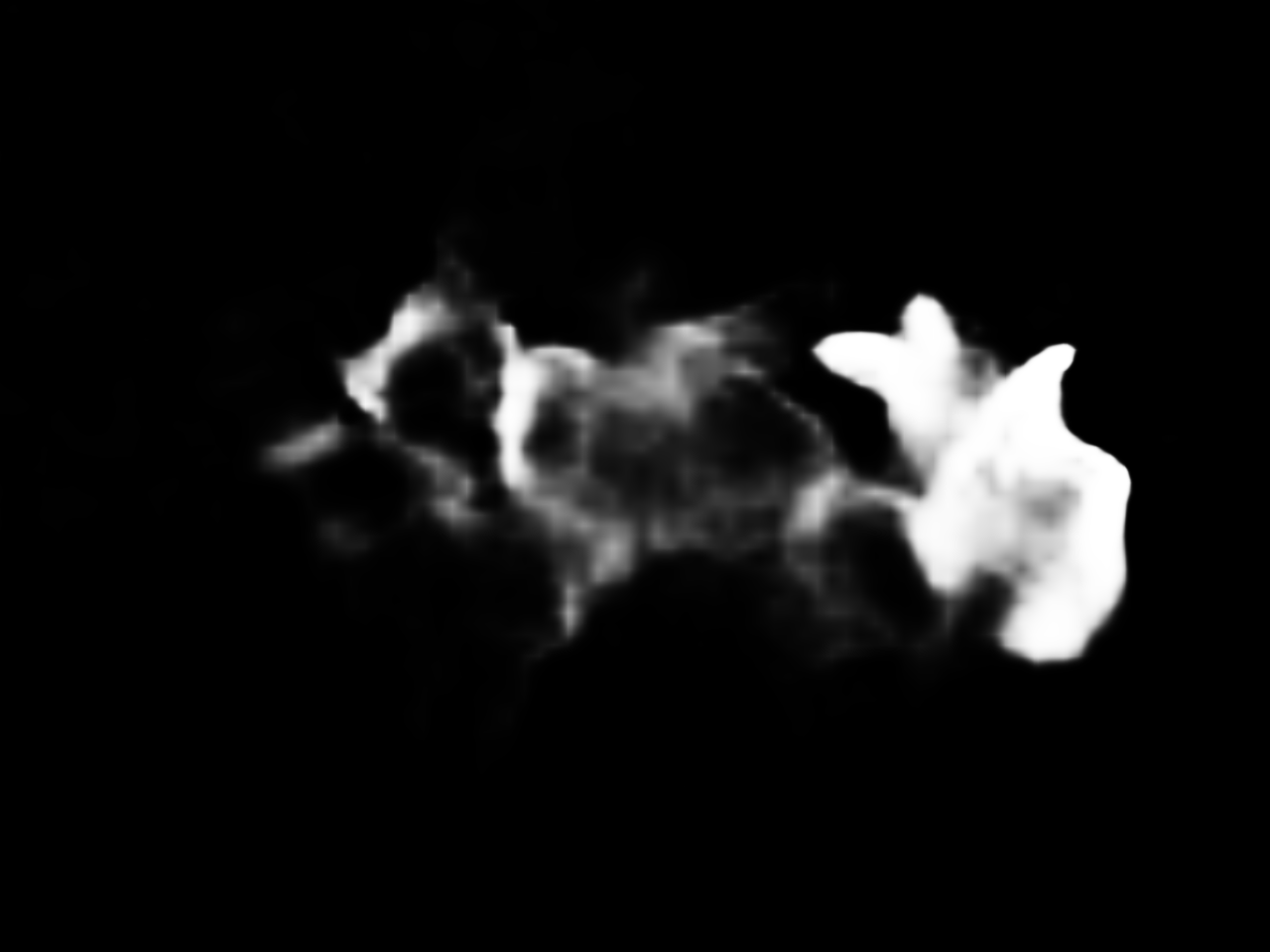}}&
  {\includegraphics[width=0.100\linewidth]{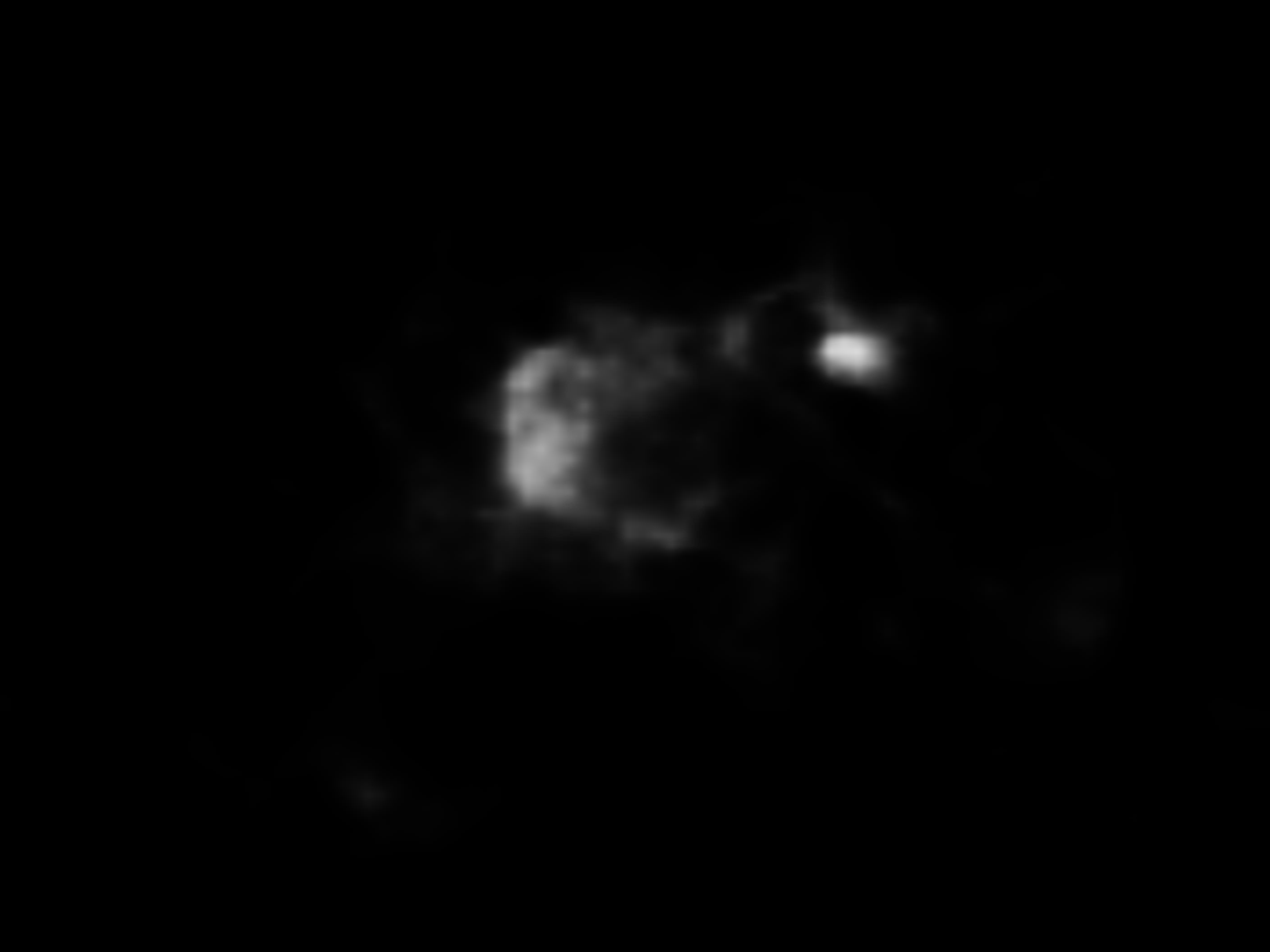}}&
    {\includegraphics[width=0.100\linewidth]{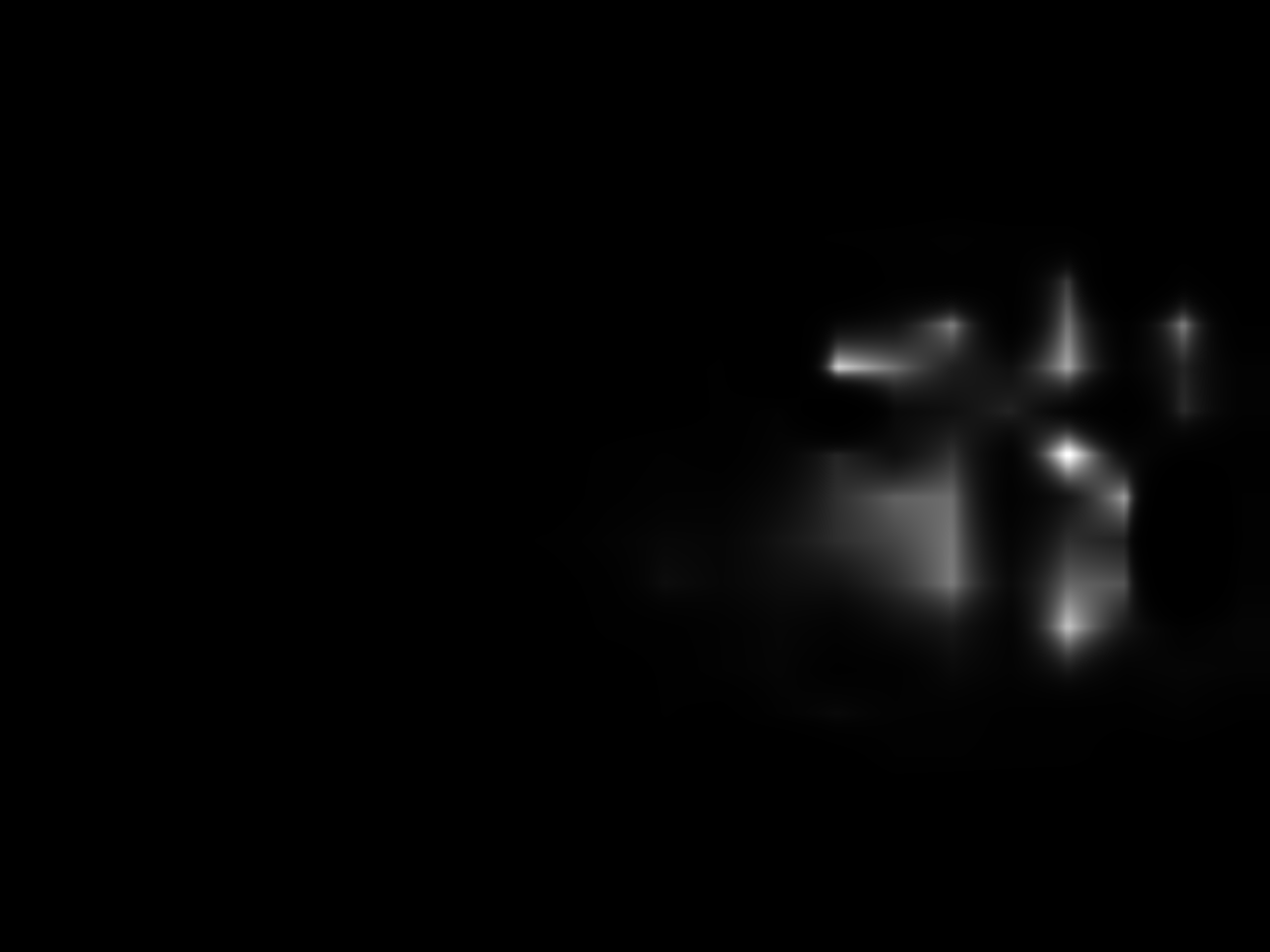}}&
    {\includegraphics[width=0.100\linewidth]{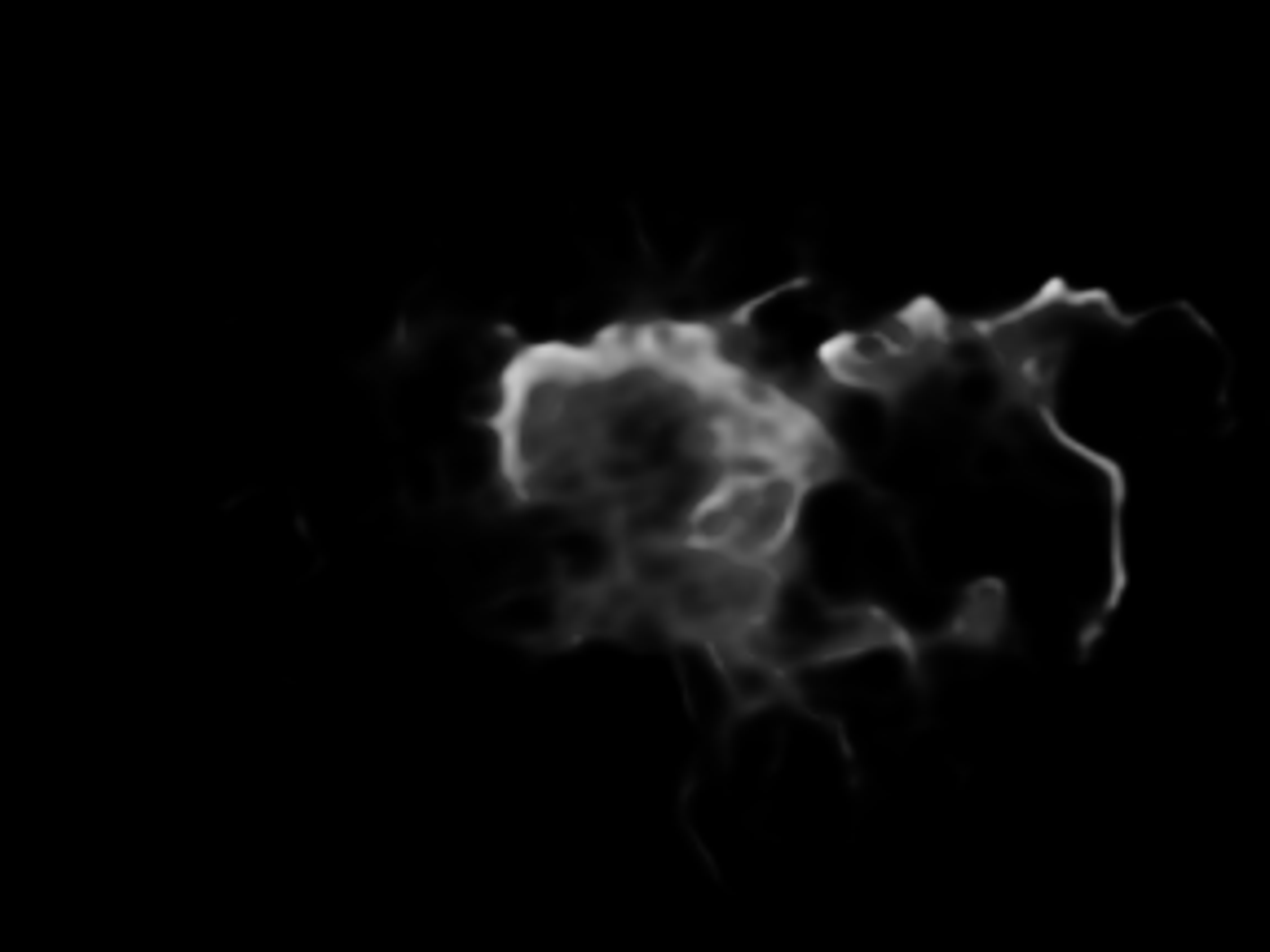}}&
    {\includegraphics[width=0.100\linewidth]{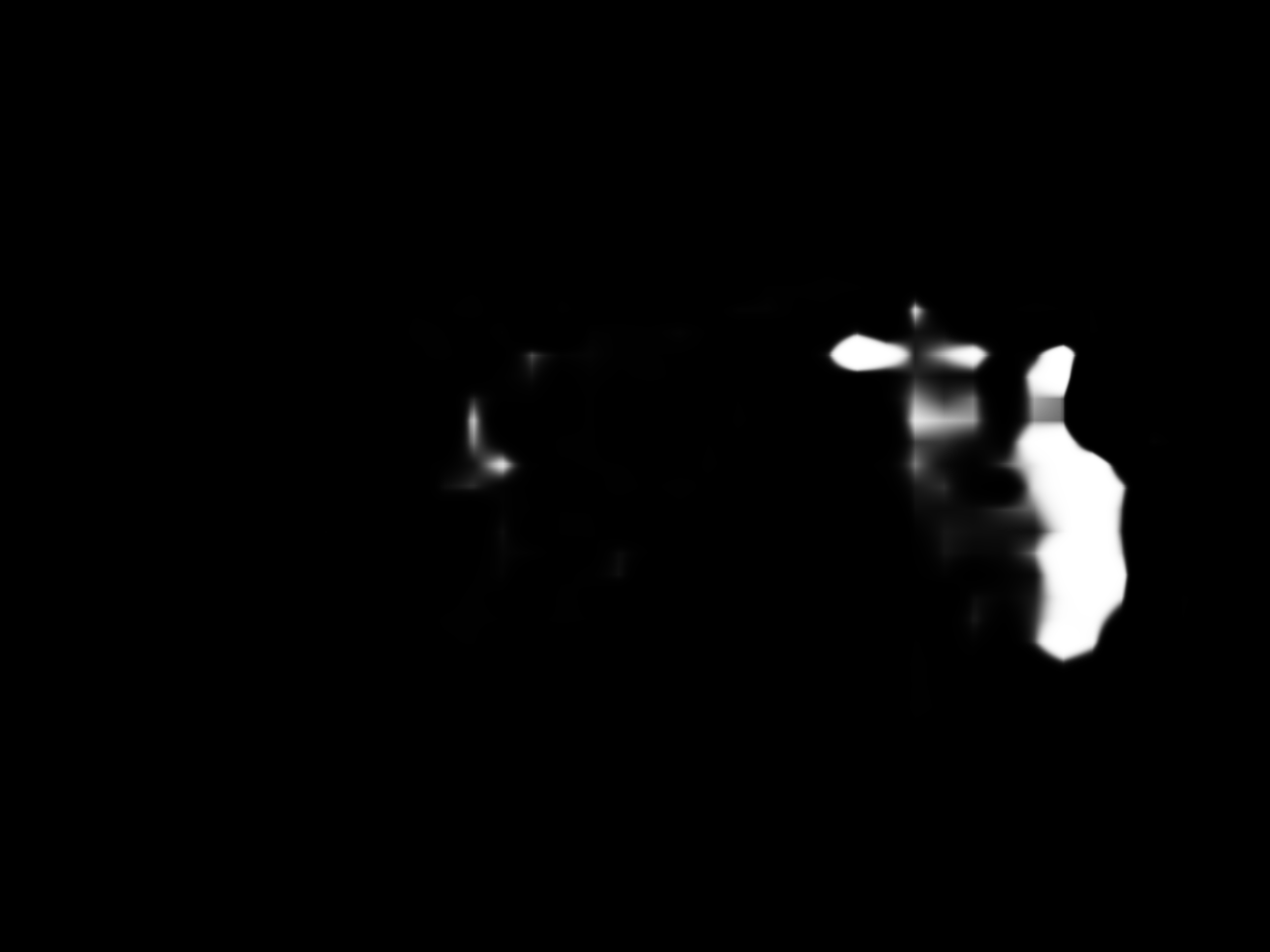}}&
    {\includegraphics[width=0.100\linewidth]{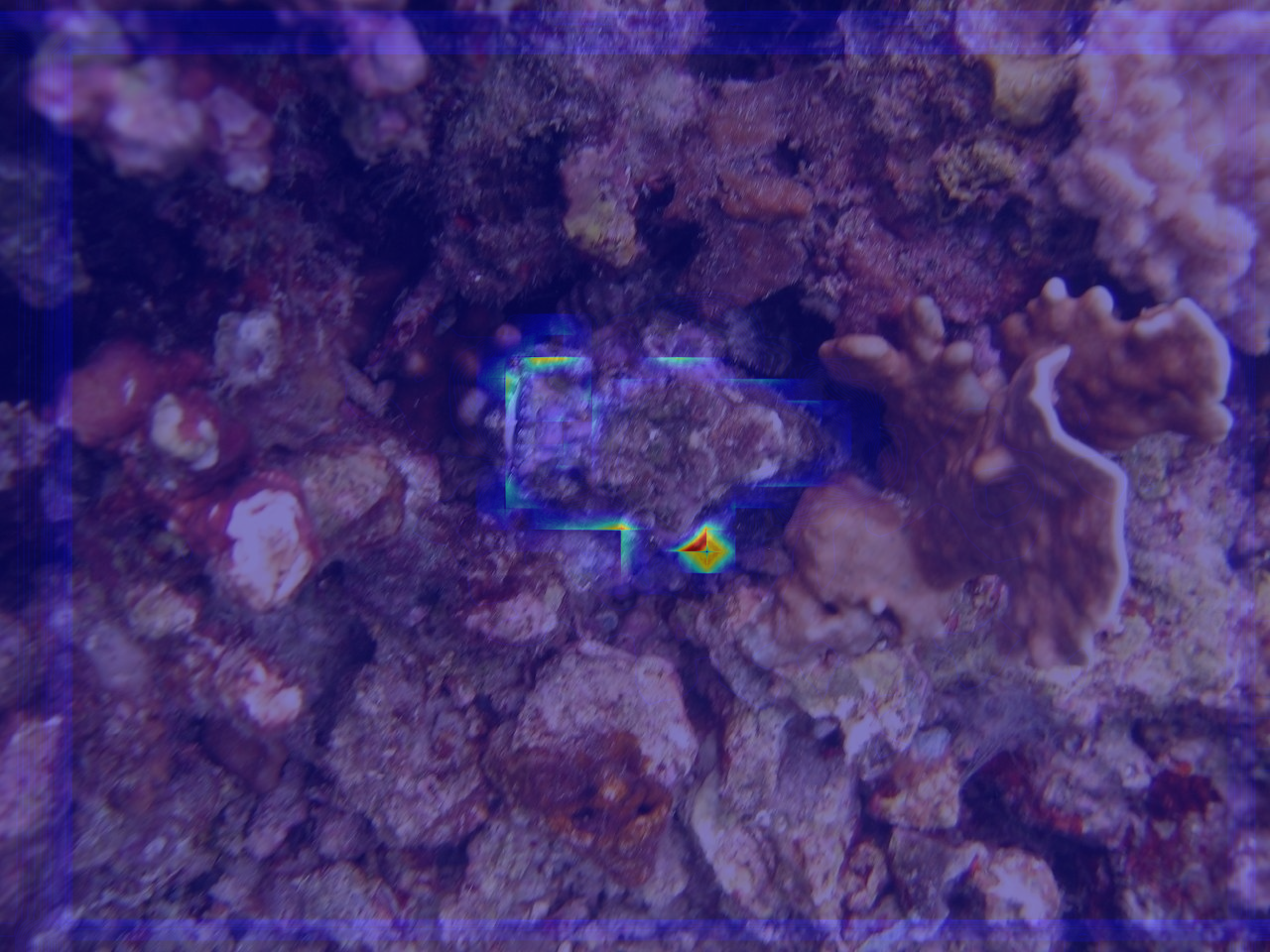}}&
    {\includegraphics[width=0.100\linewidth]{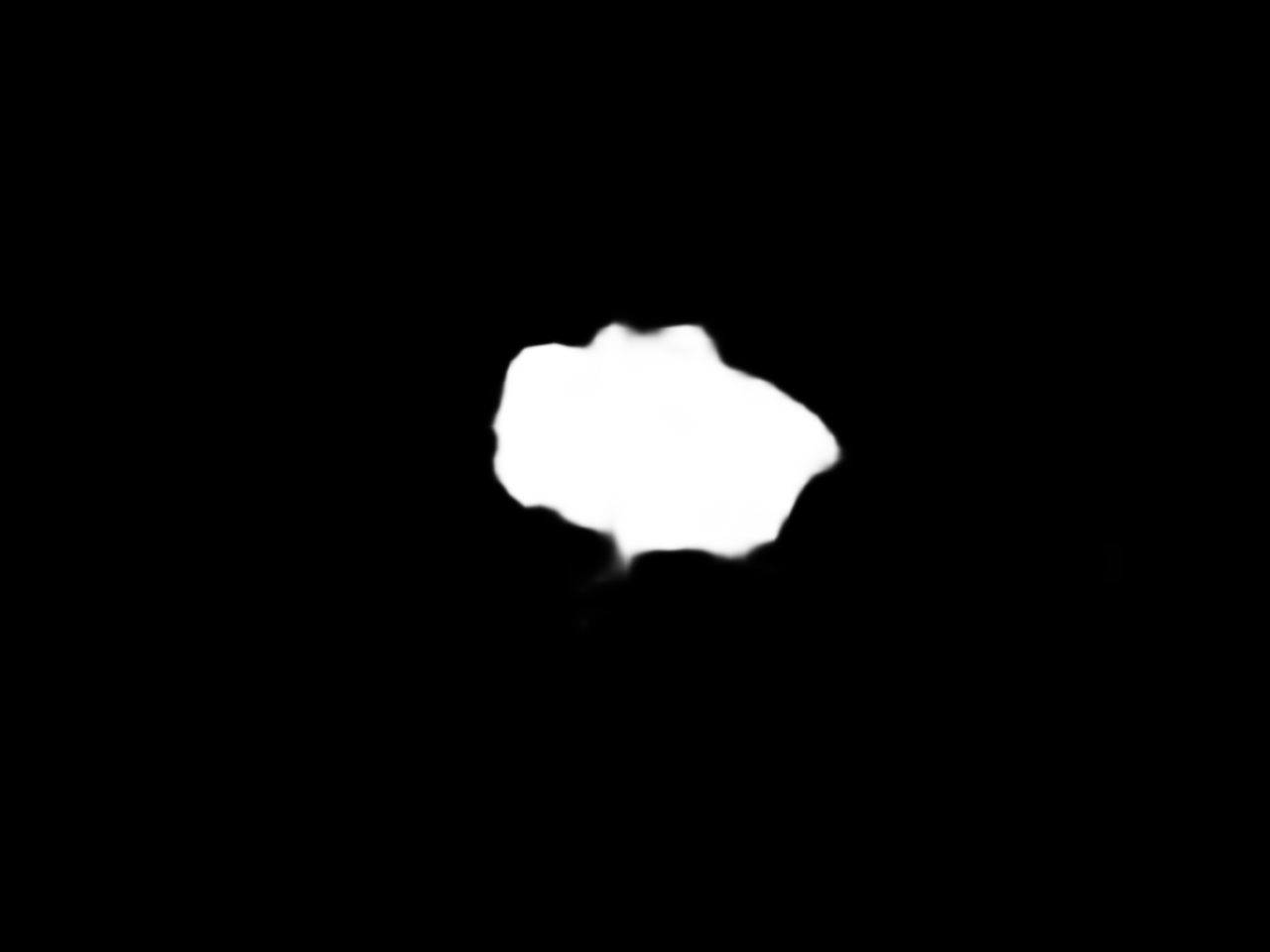}}\\
    \footnotesize{Image} &
     \footnotesize{GT} &\footnotesize{SCRN \cite{scrn_sal}} &\footnotesize{CSNet \cite{gao2020highly}}&\footnotesize{F3Net \cite{wei2020f3net}}&\footnotesize{ITSD \cite{zhou2020interactive}} &\footnotesize{SINet \cite{fan2020camouflaged}} &\footnotesize{Uncertainty} &\footnotesize{Ours}\\
   \end{tabular}
   \end{center}
    \caption{Predictions of competing camouflaged object detection models and ours.
    } 
    \label{fig:cod_samples_comparison}
\end{figure*}

\section{Experimental Results}
\subsection{Setting:}
\noindent\textbf{Dataset:} For salient object detection, we train our model using the augmented DUTS training dataset \cite{wang2017learning}, and testing on six other testing dataset, including  the DUTS testing dataset, ECSSD \cite{yan2013hierarchical}, DUT \cite{Manifold-Ranking:CVPR-2013}, HKU-IS \cite{li2015visual}, THUR \cite{THUR} and SOC testing dataset \cite{fan2018SOC}. For camouflaged object detection, we train our model using COD10K training set \cite{fan2020camouflaged}, and test on three camouflaged object detection testing sets, including CAMO \cite{le2019anabranch}, CHAMELEON \cite{Chameleon2018}, and COD10K.

\noindent\textbf{Evaluation Metrics:} We use four evaluation metrics to evaluate the performance of both SOD and COD models, including Mean Absolute Error, Mean F-measure, Mean E-measure \cite{fan2018enhanced} and S-measure \cite{fan2017structure}.

\noindent\textbf{Training details:}
We train our model in Pytorch with ResNet50 \cite{he2016deep} as backbone as shown in Fig.~\ref{fig:network_overview}. Both the encoders for saliency and camouflage branches are initialized with ResNet50 \cite{he2016deep} trained on ImageNet, and other newly added layers are randomly initialized. We resize all the images and ground truth to $352\times352$. The maximum iteration is 36000, and we iteratively update 3 times the saliency branch and then one time the camouflage branch. The initial learning rate is 2.5e-5. We adopt the \enquote{step} learning rate decay policy, and set the decay step as 24000 iteration, and decay rate as 0.1. The whole training takes 8 hours with batch size 15 on an NVIDIA GeForce RTX 2080 GPU.

\subsection{SOD performance comparison}
We compare performance of our SOD branch with eleven SOTA SOD models as shown in Table~\ref{tab:benchmark_sod_model_comparison}. One observation from Table~\ref{tab:benchmark_sod_model_comparison} is that the structure-preserving strategy is widely used in the state-of-the-art saliency detection models, \eg SCRN \cite{scrn_sal}, F$^3$Net \cite{wei2020f3net}, ITSD \cite{zhou2020interactive}, and it can indeed improve model performance. Table~\ref{tab:benchmark_sod_model_comparison} shows that we achieve 5/6 best performance, except on SOC testing dataset \cite{fan2018SOC}. The main reason is that there exists texture images in SOC, which may be treated as camouflaged object, thus influence our performance. We will investigate this issue further. Further, we show predictions of ours and SOTA models in Fig.~\ref{fig:sod_samples_comparison}, where the \enquote{Uncertainty} is obtained based on the prediction from the discriminator. Specifically, we define magnitude of gradient of the discriminator output as uncertainty following \cite{kendall2018multi}. Fig.~\ref{fig:sod_samples_comparison} shows that we produce both accurate prediction and reasonable uncertainty estimation, where the brighter area of the uncertainty map indicates the less confident region.

\subsection{COD performance comparison}
As there exists only one open source deep camouflaged object detection network (SINet \cite{fan2020camouflaged} in particular), we re-train existing saliency detection models with the camouflaged object detection training dataset \cite{fan2020camouflaged}, and test on the existing camouflaged object detection testing set. The performance of these models is shown in Table~\ref{tab:benchmark_cod_model_comparison} and Fig.~\ref{fig:cod_samples_comparison}. The consistent best performance of our camouflage model further illustrates effectiveness of the joint learning framework. Moreover, the produced uncertainty map clearly represents model confidence towards the current prediction, leading to interpretable prediction for the downstream tasks.

\subsection{Ablation study}
We present extra experiments to fully explore our model, and show performance
in Table~\ref{tab:ablation_sod_model} and
Table~\ref{tab:ablation_cod_model}.

\noindent\textbf{Train each task separately:
}
We use the same \enquote{Feature encoder} and \enquote{Prediction decoder} in Fig.~\ref{fig:network_overview} to train the SOD network and the COD network separately, and show their performance as \enquote{ASOD}
and \enquote{SCOD}
respectively.
We also train the saliency model using the original DUTS training dataset \cite{wang2017learning}, and show its performance as \enquote{SSOD}. We notice a consistent performance improvement of \enquote{ASOD} compared with \enquote{SSOD}, which indicates the effectiveness of our contrast-level data augmentation technique. The comparable performance of \enquote{ASOD} and \enquote{SCOD} with their corresponding SOTA models further prove superior performance of our new network structure.



\begin{table*}[t!]
  \centering
  \scriptsize
  \renewcommand{\arraystretch}{1.05}
  \renewcommand{\tabcolsep}{0.3mm}
  \caption{Ablation study on the salient object detection datasets.
  }
  \begin{tabular}{l|cccc|cccc|cccc|cccc|cccc|cccc}
  \hline
  &\multicolumn{4}{c|}{DUTS}&\multicolumn{4}{c|}{ECSSD}&\multicolumn{4}{c|}{DUT}&\multicolumn{4}{c|}{HKU-IS}&\multicolumn{4}{c|}{THUR}&\multicolumn{4}{c}{SOC} \\
    Method & $S_{\alpha}\uparrow$&$F_{\beta}\uparrow$&$E_{\xi}\uparrow$&$\mathcal{M}\downarrow$& $S_{\alpha}\uparrow$&$F_{\beta}\uparrow$&$E_{\xi}\uparrow$&$\mathcal{M}\downarrow$& $S_{\alpha}\uparrow$&$F_{\beta}\uparrow$&$E_{\xi}\uparrow$&$\mathcal{M}\downarrow$& $S_{\alpha}\uparrow$&$F_{\beta}\uparrow$&$E_{\xi}\uparrow$&$\mathcal{M}\downarrow$& $S_{\alpha}\uparrow$&$F_{\beta}\uparrow$&$E_{\xi}\uparrow$&$\mathcal{M}\downarrow$& $S_{\alpha}\uparrow$&$F_{\beta}\uparrow$&$E_{\xi}\uparrow$&$\mathcal{M}\downarrow$ \\ \hline
  Ours & .899 & .866 & .937 & .032 & .933 & .935 & .960 & .030 & .850 & .782 & .884 & .051 & .931 & .924 & .867 & .026 & .849 & .774 & .872 & .065 & .845 & .374 & .856 & .092  \\ \hline 
    SSOD & .885 & .850 & .921 & .036  & .914 & .913 & .837 & .036 & .835 & .758 & .867 & .054 & .918 & .905 & .950 & .029 & .836 & .753 & .852 & .069 & .810 & .362 & .833 & .129  \\
    ASOD & .891 & .854 & .928 & .035  & .920 & .918 & .946 & .035 & .840 & .767 & .871 & .053 & .920 & .909 & .956 & .028 & .838 & .760 & .859 & .068 & .829 & .368 & .849 & .111  \\
   JSOD1 & .894 & .859 & .930 & .034  & .922 & .922 & .948 & .034 & .835 & .777 & .877 & .052 & .922 & .914 & .957 & .027 & .841 & .765 & .862 & .066 & .826 & .370 & .847 & .113  \\
   JSOD2 & .893 & .860 & .930 & .034  & .928 & .929 & .954 & .031 & .839 & .770 & .873 & .054 & .921 & .915 & .958 & .027 & .839 & .763 & .859 & .067 & .826 & .368 & .844 & .114  \\
  JSOD3 & .891 & .855 & .928 & .034  & .927 & .928 & .954 & .031 & .835 & .765 & .869 & .055 & .919 & .913 & .957 & .027 & .836 & .758 & .856 & .069  & .801 & .366 & .841 & .137  \\
   \hline 
  \end{tabular}
  \label{tab:ablation_sod_model}
\end{table*}

\begin{table*}[t!]
  \centering
  \scriptsize
  \renewcommand{\arraystretch}{1.05}
  \renewcommand{\tabcolsep}{2.3mm}
  \caption{Ablation study on the camouflaged object detection datasets.}
  \begin{tabular}{l|cccc|cccc|cccc}
  \hline
  &\multicolumn{4}{c|}{CAMO}&\multicolumn{4}{c|}{CHAMELEON}&\multicolumn{4}{c}{COD10K} \\
    Method & $S_{\alpha}\uparrow$&$F_{\beta}\uparrow$&$E_{\xi}\uparrow$&$\mathcal{M}\downarrow$& $S_{\alpha}\uparrow$&$F_{\beta}\uparrow$&$E_{\xi}\uparrow$&$\mathcal{M}\downarrow$ &  $S_{\alpha}\uparrow$ & $F_{\beta}\uparrow$ & $E_{\xi}\uparrow$ & $\mathcal{M}\downarrow$  \\
  \hline
  Ours  & 0.803 & 0.759 & 0.853 & 0.076 & 0.894 & 0.848 & 0.943 & 0.030 & 0.817 & 0.726 & 0.892 & 0.035    \\ 
   \hline
   SCOD & 0.792 & 0.740 & 0.839 & 0.082  & 0.870 & 0.815 & 0.924 & 0.039 & 0.800 & 0.697 & 0.872 & 0.041  \\
   JCOD1 & 0.797 & 0.749 & 0.846 & 0.080  & 0.881 & 0.823 & 0.933 & 0.032 & 0.803 & 0.701 & 0.874 & 0.039  \\
   JCOD2 & 0.799 & 0.758 & 0.851 & 0.076  & 0.887 & 0.838 & 0.941 & 0.031 & 0.809 & 0.718 & 0.883 & 0.036  \\
   JCOD3 & 0.793 & 0.747 & 0.850 & 0.078  & 0.887 & 0.840 & 0.943 & 0.029 & 0.807 & 0.717 & 0.885 & 0.037  \\
   \hline 

  \end{tabular}
  \label{tab:ablation_cod_model}
  \vspace{-2mm}
\end{table*}
\noindent\textbf{Joint training of SOD and COD:}
We train the \enquote{Feature encoder} and \enquote{Prediction decoder} within a joint learning pipeline. The performance is shown as \enquote{JSOD1} and \enquote{JCOD1} for the saliency detection task and camouflaged object detection task respectively. The improved performance of \enquote{JSOD1} and \enquote{JCOD1} compared with \enquote{ASOD} and \enquote{SCOD} indicates that the joint learning framework can further boost performance of each task.

\noindent\textbf{Joint training of SOD and COD with similarity measure:}
We add the task connection constraint to the joint learning framework, \eg the similarity measure module in particular, and show performance as \enquote{JSOD2} and \enquote{JCOD2} respectively. In general, we can observe improved performance, especially for the COD10K dataset \cite{fan2020camouflaged}, which verifies effectiveness of our similarity measure module.

\noindent\textbf{Uncertianty-aware joint training of SOD and COD:}
Based on the joint learning framework, we include the adversarial learning pipeline to our network, and show performance as \enquote{JSOD3} and \enquote{JCOD3}. We observe relative comparable performance with the adversarial learning framework. This mainly lies in the difficulty in training the adversarial learning branch. We provide the fully convolutional discriminator in Table~\ref{tab:discriminator_gan}, and set loss for the adversarial learning empirically. A better solution could be searching for a more effective discriminator and weight, which will be our future direction.

\subsection{Hyper-parameters analysis}
In our joint learning framework, we have several hyper-parameters that influence our performance, including the maximum iteration, the interval to iteratively train the saliency and camouflage branch, the base learning rate, the dimension of the latent space, the weight of both adversarial loss and latent loss.
We found that due to the different dataset sizes and convergence rates,
the maximum iteration has a great impact on the COD task.
The SOD training dataset contains 10,553 images, which is
2.5 times of the COD dataset (the training dataset size of COD is 4,040).
To avoid overfitting on COD,
we iteratively update three times the saliency branch and then one time the camouflage branch.
For the similarity measure module, we find that the PASCAL VOC 2007 dataset \cite{everingham2007pascal} contains some samples that are both salient and camouflaged, which is contradicting with the goal of similarity measure. Thus, we train the similarity measure models every 400 iterations, and
set the weight of latent loss as 0.1.
For the \enquote{Confidence estimation} module, we observe that a large weight of the adversarial loss in Eq.~\eqref{sal_update} and Eq.~\eqref{cod_update}, \eg $\lambda_1=0.5$, may destroy the prediction, especially for the SOD branch. Our main goal of using the adversarial learning is to provide reasonable uncertainty estimation. In this case, we set the adversarial loss weight as a relative small number, \eg 0.01 in this paper, to achieve trade-off between model performance and effective uncertainty estimation.



\section{Conclusion}
In this paper, we have proposed the first joint salient object detection and camouflaged object detection network within an uncertainty-aware framework. First, we showed that the easy samples in COD dataset could be used as hard samples for SOD to learn robust SOD model. Second, by considering the contradicting attributes of these two tasks, we presented a similarity measure module to explicitly build the task connection with the extra connection modeling dataset. Lastly, we presented an adversarial learning network to explicitly model the confidence of network predictions to address the uncertainty in SOD and COD annotations. Experimental results on six benchmark SOD datasets and three benchmark COD datasets demonstrate the effectiveness of our joint learning solution.

 \section*{Acknowledgements}
\footnotesize{This research was supported in part by National Natural Science Foundation of China (61871325), National Key Research and Development Program of China (2018AAA0102803), CSIRO's Machine Learning and Artificial Intelligence Future Science Platform (MLAI FSP), and the Swiss National Science Foundation via the Sinergia grant  CRSII5-180359. We would like to thank the anonymous reviewers for their useful feedbacks.}

{\small
\bibliographystyle{ieee_fullname}
\bibliography{Camera_Ready}
}

\end{document}